\documentclass[pmlr]{jmlr}

\RequirePackage{graphicx}
 \usepackage{booktabs}
 \usepackage{amsmath}
\usepackage{amssymb}
\usepackage{mathtools}
\usepackage{multirow}
\usepackage{makecell}
\usepackage{mathrsfs}
\usepackage{microtype}
\usepackage{bm}
\usepackage{amsfonts}
\usepackage{diagbox}
\usepackage{mathtools}

\usepackage{longtable}
\makeatletter
\def\set@curr@file#1{\def\@curr@file{#1}} 
\makeatother
\usepackage[load-configurations=version-1]{siunitx} 

\theorembodyfont{\upshape}
\theoremheaderfont{\scshape}
\theorempostheader{:}
\theoremsep{\newline}

\jmlrvolume{252}
\jmlryear{2024}
\jmlrworkshop{Machine Learning for Healthcare}

\title[Fairness-aware Predictions with Contrastive Learning in Multimodal EHRs]{FairEHR-CLP: Towards Fairness-Aware Clinical \\Predictions with Contrastive Learning in \\ Multimodal Electronic Health Records}

\author{\Name{Yuqing Wang}
       \Email{ywang216@stanford.edu}\\ 
       \addr Stanford University\\
       \AND
       \Name{Malvika Pillai}
       \Email{mpillai@stanford.edu}\\
       \addr Stanford University\\
       \AND
       \Name{Yun Zhao}
       \Email{yunzhao20@meta.com}\\ 
       \addr Meta Platforms, Inc.\\
       \AND
       \Name{Catherine Curtin}
       \Email{ccurtin@stanford.edu}\\
       \addr Stanford University \\
       \AND
       \Name{Tina Hernandez-Boussard}
       \Email{boussard@stanford.edu}\\
       \addr Stanford University \\
       } 

\begin{document}

\maketitle

\begin{abstract}
In the high-stakes realm of healthcare, ensuring fairness in predictive models is crucial. Electronic Health Records (EHRs) have become integral to medical decision-making, yet existing methods for enhancing model fairness restrict themselves to unimodal data and fail to address the multifaceted social biases intertwined with demographic factors in EHRs. To mitigate these biases, we present \textit{FairEHR-CLP}: a general framework for \textbf{Fair}ness-aware Clinical \textbf{P}redictions with \textbf{C}ontrastive \textbf{L}earning in \textbf{EHR}s. FairEHR-CLP operates through a two-stage process, utilizing patient demographics, longitudinal data, and clinical notes. First, synthetic counterparts are generated for each patient, allowing for diverse demographic identities while preserving essential health information. Second, fairness-aware predictions employ contrastive learning to align patient representations across sensitive attributes, jointly optimized with an MLP classifier with a softmax layer for clinical classification tasks. Acknowledging the unique challenges in EHRs, such as varying group sizes and class imbalance, we introduce a novel fairness metric to effectively measure error rate disparities across subgroups.  Extensive experiments on three diverse EHR datasets on three tasks demonstrate the effectiveness of FairEHR-CLP in terms of fairness and utility compared with competitive baselines. FairEHR-CLP represents an advancement towards ensuring both accuracy and equity in predictive healthcare models. Our code is available at \url{https://github.com/EternityYW/FairEHR-CLP}.
\end{abstract}

\section{Introduction}

The growing availability of Electronic Health Records (EHRs) holds significant potential for enhancing healthcare delivery and patient outcomes~\citep{zhao2021bertsurv, wang2022predicting}. However, their use in predictive modeling raises substantial challenges, particularly in ensuring algorithmic fairness and addressing inherent data biases~\citep{chen2023algorithmic, giovanola2023beyond}. EHR data often mirror social and systemic biases, which if unaddressed, can perpetuate inequalities in healthcare outcomes. For example, studies have shown racial disparities in healthcare, such as Black patients being 40\% less likely to receive pain medication than White patients for similar conditions~\citep{lee2019racial}. Such biases, when ingrained in training data, can lead models to perpetuate or even exacerbate these inequalities, resulting in disparities in patient care based on race, gender, or socioeconomic status. In a field where decisions can have life-altering consequences, it is crucial to ensure that predictive tools do not inadvertently disadvantage marginalized patient groups~\citep{vela2022eliminating}. Therefore, developing fair and effective predictive models is essential.
\vspace{-0.3cm}
\begin{figure}[h]
\centering
\includegraphics[width=.8\textwidth]{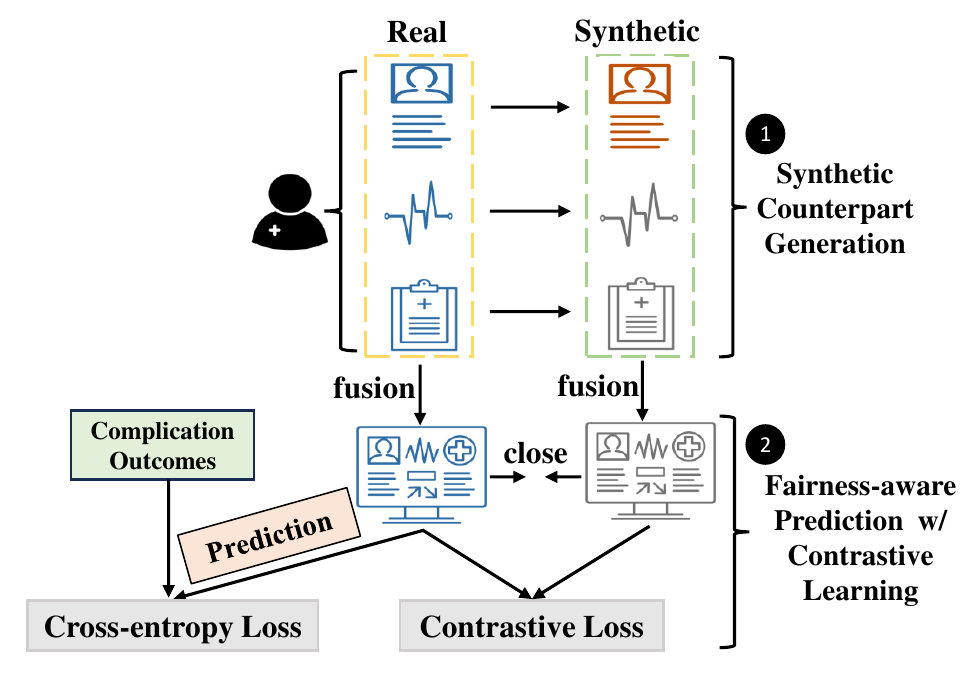}
\vspace{-0.3cm}
\caption{Overview of our FairEHR-CLP framework.}
\label{fig: model_overview}
\end{figure}

\vspace{-0.4cm}
Existing methods to enhance fairness in EHR predictive models fall into three categories, each with respective limitations. Pre-processing techniques that alter training data distributions, such as sampling~\citep{iosifidis2018dealing} and perturbation~\citep{wang2022fairness} can lead to overfitting or data distortion. Post-processing methods, involving modifications after training~\citep{du2021fairness} or prediction relabeling~\citep{lohia2019bias}, are slow and resource-intensive. In-processing strategies like loss function regularization~\citep{kim2018fairness} and adversarial training~\citep{yang2023adversarial}, overlook the interplay and complex nature of social biases~\citep{wang2021empirical, boyd2023potential, rajendran2024learning}. The intricacy involved in these techniques highlights a fundamental question: \textit{How can we develop a fair prediction model that effectively addresses the varied social biases from demographic factors in EHRs?}

To address this question, developing a fair prediction model that utilizes the value of demographic data as predictors while minimizing associated social biases is essential. Consider a scenario where a model assesses patients with similar health issues but from varied demographics, such as two individuals with cardiovascular symptoms, differing in gender and ethnicity. By applying contrastive analysis to these cases, the model can identify clinical patterns that span across demographic lines, focusing on health similarities. This strategy strengthens the model's ability to make unbiased, clinically relevant recommendations, prioritizing health factors over demographic differences. The above process aligns with the principles of contrastive learning (CL), a prominent representation learning method that differentiates similar and dissimilar instances within an embedding space~\citep{chen2020simple, chuang2020debiased, zhang2022fairness, sun2023contrastive, ge2023hyperbolic}. We aim to harness CL in balancing the use of demographics for informative predictions and the imperative for bias mitigation.

To this end, we introduce a general framework for Fairness-aware Clinical Predictions with Contrastive Learning in EHRs, which we call \textit{FairEHR-CLP}. The framework involves two distinct stages: first, synthetic counterpart generation creates synthetic instances for each patient, representing varied demographics while preserving vital health data. The second stage involves fairness-aware predictions using CL, which aims to minimize the representation distance between real patients and their synthetic counterparts who share similar health conditions but differ demographically, in tandem with a multi-layer perceptron (MLP) classifier equipped with a softmax layer for downstream classification tasks. Figure~\ref{fig: model_overview} presents an overview of our FairEHR-CLP framework.

In our experiments, we incorporate patient demographics, longitudinal data, and clinical notes into the FairEHR-CLP framework for clinical predictions. We focus on five sensitive attributes linked to social biases: gender, race, ethnicity, age, and socioeconomic status (represented by insurance type). We demonstrate the effectiveness of our method across three diverse EHR datasets: STARR~\citep{sun2021characterizing}, MIMIC-III~\citep{johnson2016mimic}, and MIMIC-IV~\citep{johnson2023mimic}, focusing on surgical patient outcomes, which are often subject to social bias~\citep{raso2023social}. We consider three binary classification tasks, identifying delirium, opioid use disorder (OUD), and 30-day readmission, all of which have a direct impact on postoperative care. Our extensive experiments show that FairEHR-CLP not only outperforms existing debiasing methods in terms of fairness but also achieves competitive predictive performance when compared to standard classification baselines.

To summarize, our contributions are three-fold:
\begin{enumerate}
\item[(1)] We develop FairEHR-CLP, a general fairness-aware clinical prediction framework that employs contrastive learning in multimodal EHRs, aiming at mitigating social biases arising from demographic factors.

\item[(2)] We propose a new fairness metric, the Error Distribution Disparity Index (EDDI), by quantifying the deviation in error rates for each subgroup from the overall error rate, particularly relevant in clinical settings with diverse group sizes and class imbalance.

\item[(3)] Extensive experiments on three large-scale EHR datasets across three classification tasks illustrate the effectiveness of our proposed method in terms of fairness and utility compared with multiple baselines.
\end{enumerate}

\subsection*{Generalizable Insights about Machine Learning in the Context of Healthcare}
Healthcare data often contain biases due to incomplete records, inaccuracies, inconsistencies, and an overrepresentation of individuals with structural privileges. These biases can be perpetuated by machine learning models, leading to disparities in predictions across different demographic groups. Additionally, healthcare data encompass a range of modalities like clinical notes, lab measurements, and demographic details. Leveraging these diverse data types can enhance model fairness by providing a more complete representation of patients. Our study underscores the critical importance of fairness in healthcare predictive models by proposing a framework, FairEHR-CLP, for fairness-aware clinical predictions that employs contrastive learning (CL) in multimodal EHRs. This method effectively reduces social biases related to demographics in the data through fairness evaluation while minimizing performance loss. The CL-based framework can be applied across clinical domains and is scalable to other types of biases in EHR data, offering a robust solution for fair machine learning applications in healthcare.

\section{Related Work}
In this section, we explore existing methods to mitigate bias and enhance fairness in EHRs, review CL applications in EHRs, and discuss fairness evaluation approaches.

\noindent \textbf{Bias and Fairness.}
EHRs, rich in patient data, often exhibit systemic biases, stemming from demographic, socioeconomic, and access disparities~\citep{zhao2021empirical, chin2023guiding, wang2023large, rajendran2024learning}. Such biases in EHRs risk being reinforced or exacerbated by algorithms trained on these datasets, potentially harming underrepresented groups. To combat this, recent research has focused on reducing algorithmic bias. Representative approaches include adversarial training~\citep{yang2023adversarial}, which involves parallel training of a task-specific classifier and a bias-exploiting adversary model, and using stacked denoising autoencoders with weighted reconstruction loss to enhance representation of underrepresented classes~\citep{sivarajkumar2023fair}. However, these approaches fail to account for the complex interactions between social biases that are embedded in demographic features and the multimodal nature of EHR data~\citep{wang2022integrating}. In contrast, our proposed method leverages multimodal EHRs and addresses a spectrum of social biases through a unified framework.

\noindent \textbf{Contrastive Learning.}
Contrastive learning (CL), originally developed for vision tasks, employs the principle of contrasting samples to identify attributes common to and differentiating between data classes~\citep{khosla2020supervised, chen2020simple, jaiswal2020survey}. In essence, CL generates varied views of original data through random augmentation, treating views from the same source as positive pairs. The model then learns effective representations by minimizing the distance between these positive pair representations. Recently, CL has been adapted for patient representation in EHRs, applied in critical event prediction for COVID-19~\citep{wanyan2021contrastive}, clinical risk prediction~\citep{zang2021scehr}, and survival analysis~\citep{nayebi2023contrastive}. However, existing CL applications in EHRs neglect potential fairness issues. To address this oversight, our method introduces a fairness-oriented contrastive loss for training models that learn fair representations, incorporating tailored contrasting sample designs specific to EHRs.

\noindent \textbf{Fairness Evaluation.}
Traditional fairness metrics such as equalized odds, equal opportunity~\citep{hardt2016equality}, demographic parity~\citep{jiang2022generalized}, and disparate impact assess fairness are based on aggregate outcomes across diverse demographic groups~\citep{feldman2015certifying}. However, these metrics may not fully capture the heterogeneity and distinct distribution patterns in EHR data, particularly when considering variability in subgroup sizes. To address this gap, we propose the Error Distribution Disparity Index (EDDI), a metric specifically designed for EHRs. EDDI measures fairness by evaluating the disparities in error rates across subgroups relative to the overall error rate, which is crucial in clinical settings characterized by imbalanced outcome labels and varying patient group sizes.

\section{Methods}
In this section, we begin by presenting an overview of the problem formulation and the workflow of our FairEHR-CLP. Then, the process of generating synthetic counterparts for each patient during the training phase is detailed. Finally, we discuss fairness-aware predictions with CL, in conjunction with the outcome prediction with the MLP classifier.

\subsection{Problem Formulation and Method Overview} \label{method_overview}
We define a dataset as $\mathscr{D} = \{ (x_k, y_k, s_k) \}_{k=1}^n$, where $x_k \in \mathcal{X}$ corresponds to the input features extracted from patient demographics, longitudinal health records, and clinical notes; $y_k \in \{0, 1\} \subseteq \mathcal{Y}$ denotes the binary target label; and $s_k \in \mathcal{S}$ signifies the sensitive attribute indicative of potential social bias. These attributes encompass gender (male, female), race (White, Black, Asian, etc.), ethnicity (including categories such as Latino/Hispanic), age (categorized into ranges like 50-60, 60-70, etc.), and socioeconomic status (SES), represented by the type of insurance (private, government, etc.). The inclusion of insurance type as a proxy for SES allows for the examination of disparities that may arise due to economic barriers to healthcare access~\citep{green2021impact}. Our objective is to develop an effective and fair prediction model $f: \mathcal{X} \rightarrow \mathcal{Y}$ that aims to accurately predict outcomes without discriminating against the subgroups defined by sensitive attributes $\mathcal{S}$ from demographics. 

Our approach unfolds in two primary stages: 1) \textbf{Synthetic Counterpart Generation}, where we generate synthetic demographic counterparts to represent a spectrum of demographic identities. For creating corresponding synthetic longitudinal data, we employ EHR-based Generative Adversarial Networks (GANs)~\citep{li2023generating}. Simultaneously, Llama2-70b~\citep{touvron2023llama} is used to synthesize clinical notes, thereby enriching our dataset to mirror demographic diversity while maintaining clinical accuracy. An example of a patient profile is illustrated in Appendix \ref{synthetic_example}. 2) \textbf{Fairness-Aware Predictions with CL}, in which we align the representations of real and synthetic data to address biases. It incorporates an MLP classifier with a softmax layer for downstream classification tasks, leveraging aligned real data representations for final prediction.

\subsection{Synthetic Counterpart Generation}
\label{synthetic}
Our initial step involves generating synthetic counterparts for sensitive attributes (i.e., gender, age, race, ethnicity, and insurance), along with longitudinal data (including vital signs and lab measurements), and clinical notes for patients during the training phase. This process creates pairs of patients with similar health conditions but distinct demographic factors. For example, if the original patient is a 55-year-old diabetic White male, the corresponding synthetic counterpart might be a 60-year-old diabetic Black female. This step enhances the representation of diverse demographics while preserving the consistency of health-related information. These synthetic samples are used alongside real data for contrastive training but not in final predictions in full FairEHR-CLP experiments.

\noindent \textbf{Sensitive Attributes.} We consider five sensitive attributes, which range from binary to multi-class subgroups for each attribute. Four of these attributes are categorical, except for age, which is a continuous variable. For the categorical variables, we randomly assign a new category to each patient to create their corresponding synthetic counterpart (e.g., male to female). For age, we segment our patient cohort into 10-year age bins (e.g., 50-60, 60-70, etc.). Then, we assign a random age within a different bin for each patient's synthetic age. 

\begin{figure}[h]
\centering
\includegraphics[width=.6\textwidth]{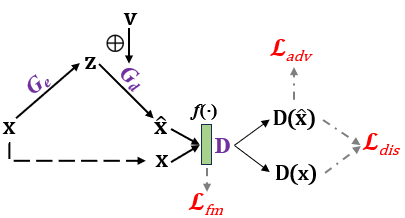}  
\caption{Architecture of EHR-GAN. The green box indicates the output from an intermediate layer of the discriminator $D$.}
\label{fig: EHR_GAN}
\end{figure}

\noindent \textbf{Longitudinal Data.}
We generate synthetic longitudinal data, which is data collected from the same individuals over a period of time, using the EHR-M-GAN model~\citep{li2023generating}, focusing exclusively on continuous data streams, which we designate as $\text{EHR-GAN}$, as shown in Figure~\ref{fig: EHR_GAN}. The architecture of $\text{EHR-GAN}$ comprises a generator $G$, which includes an encoder $G_{e}$, a decoder $G_{d}$, and a discriminator $D$. The encoder $G_{e}$ transforms the input $x$ into a latent space representation $z$. Subsequently, the decoder $G_{d}$ utilizes $z$, along with random noise $v$, to generate synthetic data $\hat{x}$. The discriminator $D$ is responsible for distinguishing between real and synthetic data. The training process involves optimizing three joint losses:
1) The discriminative loss $l_{dis}$, provided by the discriminator $D$, ensures that the generated longitudinal data appear realistic. It is defined as:
\begin{align*}
l_{dis} = -\frac{1}{n} \sum_{i=1}^n [y_i \log D(x_i) + (1-y_i) \log(1 - D(G_{d}(z_i)))],
\end{align*}
where $y_i$ denotes the label indicating whether the data is real or synthetic.
2) The adversarial loss $l_{adv}$ encourages the decoder $G_d$ to produce data that the discriminator will classify as real:
\begin{align*}
l_{adv} = -\mathbb{E}_{z \sim p_z(z)}\left[\log D(G_{d}(z))\right],
\end{align*}
where $p_z(z)$ represents the prior distribution over the latent space representation $z$. 3) The feature matching loss $l_{fm}$ ensures that the decoder $G_d$ creates data with statistical properties that are similar to real data:
\begin{align*}
l_{fm} = \sqrt{\mathbb{E}_{x \sim p_{x}(x), z \sim p_z(z)}\left[\left(f(D(x)) - f(D(G_{d}(z)))\right)^2\right]},
\end{align*}
thereby minimizing the discrepancy between the discriminative features of the real and synthetic data. Here, $f(\cdot)$ denotes the output of an intermediate layer of the discriminator $D$, and $p_{x}(x)$ is the distribution of the real data.

The total loss is $\beta_0 l_{dis} + \beta_1 l_{adv} + \beta_2 l_{fm}$, where $\beta_0, \beta_1$, and $\beta_2$ are the weighting coefficients that balance the importance of each loss component.

\noindent \textbf{Clinical Notes.}
We utilize Llama2-70b-chat~\citep{touvron2023llama} to generate synthetic clinical notes. The model receives specific instructions to ensure the preservation of essential elements in clinical documentation: “\textit{Please paraphrase the provided clinical notes, ensuring no critical medical components such as medical history, diagnoses, and treatments are omitted while maintaining the integrity of authentic documentation}”. After generation, a random subset of these synthetic notes undergoes manual review by clinical experts. This process is crucial to confirm the fidelity and accuracy of the content, ensuring it aligns with authentic clinical records in accordance with the given prompt. Details regarding the review guidelines are in Appendix~\ref{notes_guideline}.

\begin{figure*}[h]
\centering
\includegraphics[width=\textwidth]{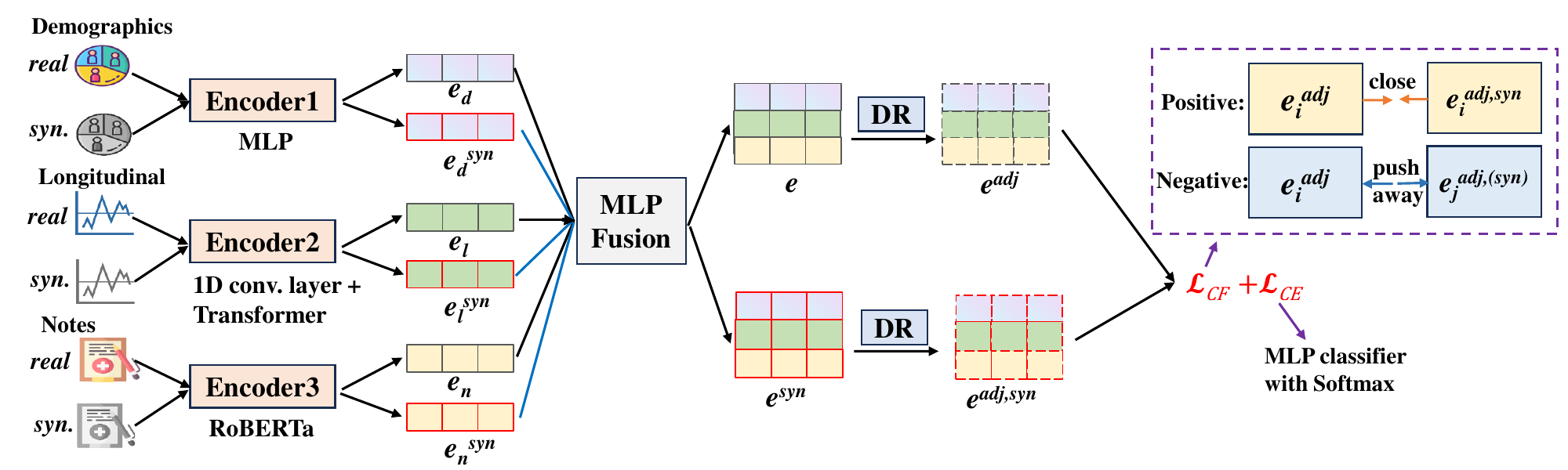}
\caption{Workflow of our second stage: fairness-aware predictions with contrastive learning. The term $e_j^{adj,(syn)}$ represents both $e_j^{adj}$ and $e_j^{adj,syn}$, with $j \neq i$. Also, conv. layer denotes the convolutional layer.}
\label{fig: cl_classification_workflow}
\end{figure*}

\subsection{Fairness-Aware Predictions with Contrastive Learning}
\label{cl}
In our augmented dataset, which includes both real patient data and synthetic counterparts spanning demographics, longitudinal records, and clinical notes during the training phase, we implement fairness-aware predictions with contrastive learning. For each patient, positive samples ($x^{+}$) are defined as their respective synthetic counterparts. These counterparts differ in sensitive demographic attributes but are matched to share similar health conditions, as determined by corresponding synthetic longitudinal and note data. In contrast, negative samples ($x^{-}$) are all other patients present in the same minibatch during training. To encode features from both real and synthetic data during training, demographic characteristics are processed using an MLP encoder, while longitudinal data are handled with a convolutional layer followed by a standard Transformer encoder to capture temporal dynamics. Clinical note embeddings are derived using RoBERTa-large~\citep{liu2019roberta}. The encoded demographic, longitudinal, and note data are denoted as $e_d$, $e_l$, and $e_n$, respectively. Following this, an MLP-based fusion combines these modality-specific representations into a unified representation that captures inter-modal dependencies and interactions: $F_{\text{fusion}} (\cdot) = \text{MLP} (e_{\text{d}} \oplus e_{\text{l}} \oplus e_{\text{n}}; \theta_{\text{fusion}})$, where $\theta_{\text{fusion}}$ represents the set of trainable parameters within the fusion layer. Integrated representations for real and synthetic data are labeled as $e$ and $e^{syn}$, respectively. To dynamically address potential biases across different data types, we introduce a Dynamic Relevance (DR) layer, defined as $F_{\text{DR}} (e) = \sigma (w) \odot e$, using $e$ as an example, where $w$ represents adjustable weights and $\sigma$ is the sigmoid function. This gating mechanism modulates the influence of each feature in the final representation. Post-DR, the adjusted embeddings are referred to as $e^{adj}$ and $e^{adj, syn}$ for real and synthetic data, respectively. The joint learning objective combines a fairness-oriented contrastive loss ($l_{CF}$) for bias mitigation and cross entropy loss ($l_{CE}$) to enhance classification performance. Formally,
\begin{align*}
l_{CF} = \sum_{k=1}^{N} -\log \frac{\exp(\text{sim}(e^{adj}_{k}, e^{adj, syn}_{k^+})/\tau)}{\sum_{j=1}^{N} \exp(\text{sim}(e^{adj}_{k}, e^{adj, syn}_{j^-})/\tau)} + \gamma \left( \frac{1}{N} \sum_{k=1}^{N} \left\| e^{adj, syn}_{k} - \mu_{\text{syn}}^{adj} \right\|_2^2 \right),
\end{align*}
where $N$ denotes the number of real embeddings (and corresponding synthetic counterparts) in a minibatch, $\text{sim} (u, v)$ calculates cosine similarity, $\tau$ is a temperature parameter, $\gamma$ is a regularization parameter, and $\mu_{\text{syn}}^{adj}$ is the mean of $e^{adj, syn}$ across a minibatch. The first term is inspired by NT-Xent loss~\citep{chen2020simple}, while the second term encourages the synthetic embeddings to cluster tightly around their mean, mitigating overfitting to outliers in synthetic data. Additionally,
\begin{align*}
l_{CE}(e, y) = -\sum_{k=1}^{N} y_{k} \log(C(e^{adj}_k)),
\end{align*}
where $y_k$ corresponds to the true label for each of the $N$ real embeddings and $C(e^{adj}_{k})$ signifies the softmax probability of the predicted class. The total loss is $\sum_k (\alpha l_{CF}$ + $(1 - \alpha) l_{CE})$, with $\alpha$ balancing fairness and performance. The detailed workflow of our second stage (fairness-aware predictions with CL) is depicted in Figure~\ref{fig: cl_classification_workflow}. For clarity, all notations used throughout this section can be found in Appendix~\ref{notations}. Implementation details are in Appendix~\ref{app_implementation}.

\section{Experimental Setup}
In this section, we outline the experimental setup, including the datasets used, the baseline models for comparison, and the evaluation metrics employed.

\subsection{Datasets} 
We evaluate our proposed framework using three EHR datasets: STAnford medicine Research data Repository (STARR) from Stanford Medicine, MIMIC-III, and MIMIC-IV. Our focus is on surgical patients aged 50 years or older, a cohort often subject to social bias in medical treatments and outcomes due to age-related factors like impaired cognition. For the MIMIC-III and MIMIC-IV datasets, we specifically employ the MIMIC-III Clinical Database CareVue subset~\citep{johnson2022subset} to ensure there is no overlap of patient data. The study targets three critical tasks: classifying delirium, OUD, and 30-day readmission. These tasks are chosen for their direct impact on enhancing postoperative care, improving patient safety, and reducing healthcare costs. Demographic indicators are excluded from clinical notes to focus solely on health conditions. We extract patient data from a 24-hour postoperative period and employ MICE imputation~\citep{van2011mice} to address missing values for all datasets. Each task is approached as a binary classification problem. The class distribution for each task is summarized in Table~\ref{tab: class_dist} with more details in Appendix~\ref{app_datasets}.

\begin{table*}[ht]
  \caption{Class distribution in three prediction tasks over all datasets.}
  \vspace{0.2cm}
  \centering
  \label{tab: class_dist}
  \begin{tabular}{cccc}
    \toprule
   \multirow{2}{*}{\textbf{Dataset}} & \makecell{Delirium} &  \makecell{OUD} & \makecell{30-day Readmission}\\
   & \textit{class 0 / 1} & \textit{class 0 / 1} & \textit{class 0 / 1} \\
    \midrule
    STARR & 39,516 / 7,417 & 42,156 / 4,777 & 34,919 / 12,014 \\
    MIMIC-III & 4,030 / 272 & 3,998 / 304 & 3,974 / 328\\
    MIMIC-IV &  7,956 / 7,962 & 14,169 / 1,749 & 9,136 / 6,782 \\
  \bottomrule
\end{tabular}
\end{table*}

\vspace{-0.5cm}
\subsection{Baselines}
To assess our method in terms of performance and fairness, we compare it with a variety of established methods. Our evaluation begins with the Demographic-free Classification (DfC) approach, based on the premise that models, if unaware of demographic features often central to socially sensitive biases, should demonstrate minimal differences in performance. Additionally, we explore two notable debiasing strategies tailored for EHR: Adversarial Debiasing (AdvDebias)\citep{zhang2018mitigating, yang2023adversarial}, a technique that simultaneously trains a classifier and an adversary model to neutralize bias, and Fair Patient Model (FPM)\citep{sivarajkumar2023fair}, which employs a Stacked Denoising Autoencoder and a weighted reconstruction loss for equitable patient representations. Furthermore, we include comparisons with embedding methods RoBERTa-large~\citep{liu2019roberta} and ClinicalBERT~\citep{alsentzer2019publicly}, widely used in general and healthcare-specific applications, respectively. The embeddings generated by these models are utilized as inputs for an MLP classifier equipped with a softmax layer for prediction.

\subsection{Evaluation Metrics}
For classification performance evaluation, we employ F1 and AUROC as metrics. Regarding fairness metrics, we adopt a variant of the Equalized Odds (EO) metric~\citep{hardt2016equality}, a widely recognized notion of group fairness~\citep{dwork2012fairness}. Traditionally, EO suggests that a model achieves fairness when the True Positive Rates (TPR) and False Positive Rates (FPR) are consistent across all subgroups defined by the sensitive attribute. However, this conventional interpretation of EO may not fully account for practical challenges such as data variability or differences in group sizes in clinical settings. Therefore, we employ the Average Disparity in EO to measure the average deviation from the ideal EO condition:
\begin{align*}
\text{EO}_{\text{TPR}} &= \frac{1}{\binom{|S|}{2}} \sum_{s_i} \sum_{s_j > s_i} \left| \text{TPR}_{s_i} - \text{TPR}_{s_j} \right|, \\
\text{EO}_{\text{FPR}} &= \frac{1}{\binom{|S|}{2}} \sum_{s_i} \sum_{s_j > s_i} \left| \text{FPR}_{s_i} - \text{FPR}_{s_j} \right|,
\end{align*}
where
\begin{align*}
\text{TPR}_s = \frac{\text{TP}_s}{\text{TP}_s + \text{FN}_s}
\end{align*}
and 
\begin{align*}
\text{FPR}_s = \frac{\text{FP}_s}{\text{FP}_s + \text{TN}_s}.
\end{align*}
Here, for each subgroup $s \in S$, where $S$ is the set of subgroups determined by a sensitive attribute (e.g., race), $\text{TP}_s$, $\text{FN}_s$, $\text{FP}_s$, and $\text{TN}_s$ represent the counts of true positives, false negatives, false positives, and true negatives for each subgroup $s$, respectively. We adopt the pairwise comparison approach, averaging the differences in TPR and FPR across all pairs of subgroups (e.g., White, Black, etc.) within a sensitive attribute (e.g., race). We then compute the arithmetic mean of $\text{EO}_{\text{TPR}}$ and $\text{EO}_{\text{FPR}}$ to establish a singular EO metric. 

A critical limitation of the traditional EO metric is its tendency to oversimplify fairness across subgroups that are diverse and unevenly represented, failing to adequately capture subgroup-specific error rate disparities. To overcome this, we introduce the Error Distribution Disparity Index (EDDI), a new fairness metric designed to address the complexities of clinical settings, especially those with significant data variability and diverse group sizes. It is formulated as:
\begin{align*}
\text{EDDI} &= \frac{1}{|S|} \sum_{s \in S} \frac{\text{ER}_s - \text{OER}}{\max(\text{OER}, 1 - \text{OER})},
\end{align*}
where 
\begin{align*}
\text{ER}_s = \frac{1}{N_s} \sum_{i \in s} \mathbb{I}(y_i \neq \hat{y}_i)
\end{align*} represents the error rate for each subgroup $s$ and \begin{align*}
\text{OER} = \frac{1}{N} \sum_{i=1}^{N} \mathbb{I}(y_i \neq \hat{y}_i)
\end{align*} 
denotes the overall error rate across the dataset. Here, $y_i$ and $\hat{y}_i$ denote the true and predicted labels, respectively. $N_s$ and $N$ indicate the number of instances within each subgroup and the total number of instances in the dataset, respectively. EDDI quantifies the error rate deviation for each subgroup from the overall error rate. We contend that a model is fair if it maintains consistent error rates across all demographic subgroups. In general, reduced values of EO and EDDI signify enhanced fairness in the model.

\section{Results}
In this section, we present a comprehensive comparison of our method with baselines across all datasets in Section~\ref{main_results}, explore the effects of data modalities, model components, and hyperparameters in Section~\ref{ablation_study}, provide visualizations of learned representations in Section~\ref{tsne_visualization}, and analyze the model's impact on each sensitive attribute in Section~\ref{case_study}.

\begin{table*}[t]
\centering
\caption{Performance and fairness evaluation across three Datasets: STARR, MIMIC-III, and MIMIC-IV. We report average results and standard deviations over five runs. EO and EDDI results are averaged over five sensitive attributes. For each dataset and each task, results highlighted in bold indicate the highest performance, while those underlined denote the optimal fairness outcomes. Our method demonstrates superior classification performance and fairness in the majority of settings.}
\vspace{0.1cm}
\resizebox{\linewidth}{!}{%
\label{tab: main_results}
\begin{tabular}{lcccccccccccc}
\toprule
\multirow{2}{*}{\textbf{Model}} & \multicolumn{4}{c}{\textbf{Delirium}} & \multicolumn{4}{c}{\textbf{OUD}} & \multicolumn{4}{c}{\textbf{30-Day Readmission}} \\ 
\cmidrule(lr){2-5} \cmidrule(lr){6-9} \cmidrule(lr){10-13}
 & F1 ($\uparrow$) & AUROC ($\uparrow$) & EO ($\downarrow$) & EDDI ($\downarrow$) & F1 ($\uparrow$) & AUROC ($\uparrow$) & EO ($\downarrow$) & EDDI ($\downarrow$) & F1 ($\uparrow$) & AUROC ($\uparrow$) & EO ($\downarrow$) & EDDI ($\downarrow$)\\ 
\midrule
\multicolumn{13}{c}{\textbf{Dataset 1: STARR}} \\
\midrule
DfC  &  $79.6_{\pm 1.4}$ &  $81.8_{\pm 1.2}$ &  \underline{$5.2_{\pm 0.8}$}  &  \underline{$2.6_{\pm 0.5}$}    &  $85.7_{\pm 1.8}$  &   $89.2_{\pm 1.5}$    &  \underline{$3.4_{\pm 0.6}$}  &  \underline{$2.8_{\pm 0.7}$}    &  $80.9_{\pm 1.6}$  &   $83.4_{\pm 1.5}$    &  \underline{$0.2_{\pm 0.6}$}  &  \underline{$3.7_{\pm 0.5}$}    \\
AdvDebias  & $81.5_{\pm 1.7}$ &  $83.8_{\pm 1.4}$ &  $6.6_{\pm 0.9}$  &  $4.2_{\pm 0.5}$    &  $83.6_{\pm 2.0}$  &   $87.3_{\pm 1.6}$    &  $3.8_{\pm 0.8}$  &  $2.9_{\pm 0.6}$    &  $81.2_{\pm 1.8}$  &   $84.2_{\pm 1.4}$    &  $0.8_{\pm 0.4}$  &  $4.8_{\pm 0.6}$  \\
FPM  &    $80.2_{\pm 1.7}$ &  $82.6_{\pm 1.4}$ &  $7.0_{\pm 0.9}$  &  $4.4_{\pm 0.6}$    &  $84.3_{\pm 2.1}$  &   $88.1_{\pm 1.8}$    &  $3.8_{\pm 0.9}$  &  $3.0_{\pm 0.8}$    &  $80.6_{\pm 1.2}$  &   $83.1_{\pm 1.0}$    &  $0.9_{\pm 0.3}$  &  $4.7_{\pm 0.6}$   \\
RoBERTa   &  $83.6_{\pm 1.5}$ &  $86.2_{\pm 1.3}$ &  $8.7_{\pm 1.0}$  &  $5.2_{\pm 0.8}$    &  \bm{$87.5_{\pm 1.8}$}  &   \bm{$91.3_{\pm 1.5}$}    &  $4.8_{\pm 0.7}$  &  $4.0_{\pm 0.8}$    &  $82.3_{\pm 1.7}$  &   $85.9_{\pm 1.6}$    &  $1.4_{\pm 0.3}$  &  $5.9_{\pm 0.9}$      \\
ClinicalBERT  &  $82.8_{\pm 1.6}$ &  $84.1_{\pm 1.4}$ &  $8.0_{\pm 1.1}$  &  $4.6_{\pm 0.7}$    &  $85.2_{\pm 1.4}$  &   $88.9_{\pm 1.2}$    &  $4.2_{\pm 0.8}$  &  $3.5_{\pm 0.7}$    &  $81.6_{\pm 1.9}$  &   $84.7_{\pm 1.3}$    &  $1.1_{\pm 0.4}$  &  $5.6_{\pm 0.8}$      \\
FairEHR-CLP (Ours)  &   \bm{$84.1_{\pm 1.3}$} &  \bm{$87.3_{\pm 1.0}$} &  $5.7_{\pm 0.7}$  &  $3.4_{\pm 0.5}$    &  $86.3_{\pm 1.6}$  &   $90.6_{\pm 1.4}$    &  $3.5_{\pm 0.6}$  &  \underline{$2.8_{\pm 0.5}$}    &  \bm{$83.2_{\pm 1.3}$}  &   \bm{$87.8_{\pm 1.5}$}    &  $0.4_{\pm 0.2}$  &  $4.4_{\pm 0.6}$   \\
\midrule
\multicolumn{13}{c}{\textbf{Dataset 2: MIMIC-III}} \\
\midrule
DfC  &   $82.9_{\pm 1.4}$ &  $85.8_{\pm 1.3}$ & \underline{$5.8_{\pm 0.7}$} & \underline{$3.6_{\pm 0.6}$} & $86.8_{\pm 1.5}$ & $88.3_{\pm 1.6}$ & \underline{$3.3_{\pm 0.5}$} & \underline{$1.8_{\pm 0.5}$} & $83.7_{\pm 1.2}$ & $86.6_{\pm 1.0}$ & \underline{$2.1_{\pm 0.3}$} & \underline{$1.3_{\pm 0.4}$}    \\
AdvDebias  &  $74.5_{\pm 1.6}$ &  $77.6_{\pm 1.7}$ & $7.0_{\pm 0.9}$ & $4.9_{\pm 0.8}$ & $85.2_{\pm 1.5}$ & $87.1_{\pm 1.3}$ & $3.9_{\pm 0.7}$ & $2.4_{\pm 0.5}$ & $85.4_{\pm 1.3}$ & $88.1_{\pm 1.0}$ & $5.1_{\pm 0.3}$ & $3.8_{\pm 0.2}$ \\
FPM  & $75.8_{\pm 1.6}$ &  $79.2_{\pm 1.8}$ & $6.6_{\pm 0.8}$ & $4.3_{\pm 0.6}$ & $83.7_{\pm 1.3}$ & $86.4_{\pm 1.5}$ & $4.5_{\pm 0.3}$ & $2.6_{\pm 0.4}$ & $84.3_{\pm 1.4}$ & $87.2_{\pm 1.2}$ & $5.4_{\pm 0.5}$ & $4.0_{\pm 0.4}$   \\
RoBERTa   &  $83.7_{\pm 1.4}$ &  $86.9_{\pm 1.6}$ & $7.2_{\pm 0.7}$ & $4.0_{\pm 0.6}$ & $87.2_{\pm 1.3}$ & $89.7_{\pm 1.5}$ & $4.6_{\pm 0.6}$ & $2.9_{\pm 0.6}$ & $86.1_{\pm 1.4}$ & $89.5_{\pm 1.2}$ & $5.6_{\pm 0.2}$ & $4.5_{\pm 0.3}$  \\
ClinicalBERT  & $85.1_{\pm 1.5}$ &  $87.6_{\pm 1.7}$ & $6.7_{\pm 0.8}$ & $4.2_{\pm 0.7}$ & $86.9_{\pm 1.4}$ & $88.5_{\pm 1.3}$ & $4.2_{\pm 0.5}$ & $3.5_{\pm 0.7}$ & $85.3_{\pm 1.4}$ & $87.9_{\pm 1.6}$ & $5.3_{\pm 0.6}$ & $4.8_{\pm 0.8}$  \\
FairEHR-CLP (Ours) &  \bm{$85.5_{\pm 1.2}$} &  \bm{$89.7_{\pm 1.1}$} & $6.2_{\pm 0.3}$ & $3.8_{\pm 0.5}$ & \bm{$89.4_{\pm 1.4}$} & \bm{$91.9_{\pm 1.5}$} & $3.7_{\pm 0.5}$ & $2.0_{\pm 0.4}$ & \bm{$88.2_{\pm 1.3}$} & \bm{$91.4_{\pm 1.1}$} & $3.3_{\pm 0.4}$ & $2.1_{\pm 0.6}$ \\
\midrule
\multicolumn{13}{c}{\textbf{Dataset 3: MIMIC-IV}} \\
\midrule
DfC  &  $76.1_{\pm 1.6}$ &  $79.4_{\pm 1.3}$ & \underline{$4.9_{\pm 0.6}$} & \underline{$3.5_{\pm 0.4}$} & $75.2_{\pm 1.9}$ & $79.5_{\pm 1.8}$ & \underline{$1.3_{\pm 0.6}$} & \underline{$2.1_{\pm 0.5}$} & $76.9_{\pm 1.5}$ & $79.3_{\pm 1.4}$ & \underline{$2.2_{\pm 0.6}$} & \underline{$4.5_{\pm 0.7}$}  \\
AdvDebias  &  $73.6_{\pm 1.8}$ &  $76.6_{\pm 1.6}$ & $5.3_{\pm 0.7}$ & $4.0_{\pm 0.8}$ & $74.7_{\pm 1.5}$ & $78.6_{\pm 1.3}$ & $5.8_{\pm 0.5}$ & $3.0_{\pm 0.6}$ & $77.8_{\pm 1.3}$ & $80.6_{\pm 1.2}$ & $3.1_{\pm 0.3}$ & $5.9_{\pm 0.3}$ \\
FPM  &  $70.4_{\pm 2.0}$ &  $73.1_{\pm 1.8}$ & $5.6_{\pm 0.8}$ & $4.2_{\pm 0.9}$ & $72.9_{\pm 1.5}$ & $76.0_{\pm 1.3}$ & $5.0_{\pm 0.8}$ & $2.6_{\pm 0.7}$ & $79.2_{\pm 1.4}$ & $82.7_{\pm 1.5}$ & $3.0_{\pm 0.5}$ & $5.6_{\pm 0.7}$  \\
RoBERTa  &  $77.9_{\pm 1.4}$ &  $81.1_{\pm 1.6}$ & $5.7_{\pm 0.5}$ & $4.3_{\pm 0.7}$ & \bm{$86.3_{\pm 1.9}$} & \bm{$89.6_{\pm 1.7}$} & $4.2_{\pm 0.8}$ & $2.3_{\pm 0.9}$ & $81.3_{\pm 1.4}$ & $85.7_{\pm 1.5}$ & $3.6_{\pm 0.6}$ & $5.6_{\pm 0.5}$  \\
ClinicalBERT  &  $78.2_{\pm 1.7}$ &  $81.7_{\pm 1.5}$ & $6.0_{\pm 0.6}$ & $4.6_{\pm 0.8}$ & $84.2_{\pm 2.1}$ & $87.6_{\pm 1.8}$ & $4.9_{\pm 0.9}$ & $3.1_{\pm 0.9}$ & $80.4_{\pm 1.2}$ & $83.7_{\pm 1.1}$ & $3.9_{\pm 0.5}$ & $5.7_{\pm 0.6}$ \\
FairEHR-CLP (Ours)  &  \bm{$78.8_{\pm 1.2}$} &  \bm{$82.4_{\pm 1.0}$} & $6.1_{\pm 0.4}$ & \underline{$3.5_{\pm 0.3}$} & $84.8_{\pm 1.6}$ & $88.9_{\pm 1.5}$ & $1.5_{\pm 0.3}$ & $3.0_{\pm 0.6}$ & \bm{$81.6_{\pm 1.8}$} & \bm{$86.4_{\pm 1.6}$} & $2.8_{\pm 0.7}$ & $5.2_{\pm 0.9}$\\
\bottomrule
\end{tabular}}
\end{table*}

\subsection{Main Results}
\label{main_results}
We report the classification and fairness results from the test set in the second stage of our approach (see Figure~\ref{fig: model_overview}) across three tasks and three datasets (9 settings in total) in Table~\ref{tab: main_results}. We use F1 and AUROC as performance metrics, as well as EO and EDDI as fairness metrics, with EO and EDDI results averaged over five sensitive attributes. There are several key takeaways. Firstly, FairEHR-CLP consistently outperforms DfC in F1 and AUROC by 4.8\% and 5.8\% on average, respectively, highlighting the benefit of demographic features in enhancing predictive accuracy, despite potential bias risks. In terms of fairness, FairEHR-CLP achieves EO and EDDI levels comparable to DfC, affirming the effectiveness of our bias mitigation approach. Moreover, when compared with specialized debiasing methods like AdvDebias and FPM, FairEHR-CLP excels in both predictive accuracy and fairness in most settings. This superior performance can be attributed to its comprehensive integration of multimodal EHR data and concurrent bias mitigation across multiple sensitive attributes, in contrast to the single-attribute focus of AdvDebias and FPM. Lastly, against classification methods using embeddings such as RoBERTa and ClinicalBERT, FairEHR-CLP shows superior performance in 7 out of 9 tasks, along with consistently lower EO and EDDI scores across all settings, demonstrating its robustness in balancing bias management with minimal performance loss.

\subsection{Ablation Study}
\label{ablation_study}
We conduct ablation studies on the STARR dataset to evaluate: (1) the effectiveness of various data modalities; (2) the impact of the main components of FairEHR-CLP; and (3) the influence of the key hyperparameter $\alpha$, which balances fairness and performance. For additional results on other datasets, please refer to Appendix~\ref{app_ablation}.

\textbf{Data Modalities.}
We study the effectiveness of different data modalities (demographics $\mathcal{D}$, longitudinal $\mathcal{L}$, and notes $\mathcal{N}$) within the full FairEHR-CLP framework. Considering our objective of mitigating social bias, often rooted in $\mathcal{D}$, we keep it constant in our ablation experiments. We then explore all combinations involving $\mathcal{D}$ and present the results on the STARR dataset in Table~\ref{tab: ablation_data_results}. We observe that the $\mathcal{D + L}$ combination marginally outperforms the $\mathcal{D + N}$ combination. Utilizing the full dataset ($\mathcal{D + L + N}$) results in a 2.2\% improvement in F1 and a 2.5\% increase in AUROC compared to the second-best results ($\mathcal{D + L}$). From a fairness perspective, the complete data combination consistently demonstrates a reduction in bias, indicating a more nuanced understanding and representation of patient profiles, leading to more equitable outcome predictions.

\begin{table*}[htbp]
\centering
\caption{Effects of different data modalities as inputs for FairEHR-CLP on the STARR dataset. Here, $\mathcal{D}$, $\mathcal{L}$, and $\mathcal{N}$ represent demographics, longitudinal data, and clinical notes, respectively.}
\vspace{0.1cm}
\resizebox{\linewidth}{!}{%
\label{tab: ablation_data_results}
\begin{tabular}{ccccccccccccc}
\toprule
\multirow{2}{*}{\textbf{\makecell{Data \\ Modalities}}} & \multicolumn{4}{c}{\textbf{Delirium}} & \multicolumn{4}{c}{\textbf{OUD}} & \multicolumn{4}{c}{\textbf{30-Day Readmission}} \\ 
\cmidrule(lr){2-5} \cmidrule(lr){6-9} \cmidrule(lr){10-13}
 & F1 ($\uparrow$) & AUROC ($\uparrow$) & EO ($\downarrow$) & EDDI ($\downarrow$) & F1 ($\uparrow$) & AUROC ($\uparrow$) & EO ($\downarrow$) & EDDI ($\downarrow$) & F1 ($\uparrow$) & AUROC ($\uparrow$) & EO ($\downarrow$) & EDDI ($\downarrow$)\\ 
\midrule
$\mathcal{D}$ & $78.5_{\pm 1.7}$ & $80.2_{\pm 1.6}$ & $7.8_{\pm 0.9}$ & $5.5_{\pm 0.6}$ & $81.2_{\pm 1.5}$ & $85.0_{\pm 1.3}$ & $4.7_{\pm 0.8}$ & $4.1_{\pm 0.7}$ & $79.3_{\pm 1.4}$ & $82.1_{\pm 1.2}$ & $6.0_{\pm 0.5}$ & $6.2_{\pm 0.6}$ \\
$\mathcal{D + L}$ & $82.3_{\pm 1.4}$ & $85.5_{\pm 1.2}$ & $6.2_{\pm 0.8}$ & $4.2_{\pm 0.5}$ & $84.1_{\pm 1.6}$ & $88.3_{\pm 1.4}$ & $3.9_{\pm 0.7}$ & $3.0_{\pm 0.6}$ & $81.8_{\pm 1.3}$ & $85.4_{\pm 1.1}$ & $1.1_{\pm 0.3}$ & $4.9_{\pm 0.5}$ \\
$\mathcal{D + N}$ & $81.7_{\pm 1.5}$ & $84.8_{\pm 1.3}$ & $6.7_{\pm 0.7}$ & $4.8_{\pm 0.6}$ & $83.7_{\pm 1.7}$ & $87.6_{\pm 1.5}$ & $4.1_{\pm 0.6}$ & $3.5_{\pm 0.7}$ & $81.5_{\pm 1.2}$ & $85.2_{\pm 1.0}$ & $1.6_{\pm 0.4}$ & $5.0_{\pm 0.7}$ \\
$\mathcal{D + L + N}$  &   \bm{$84.1_{\pm 1.3}$} &  \bm{$87.3_{\pm 1.0}$} &  \underline{$5.7_{\pm 0.7}$}  &  \underline{$3.4_{\pm 0.5}$}    &  \bm{$86.3_{\pm 1.6}$}  &   \bm{$90.6_{\pm 1.4}$}    &  \underline{$3.5_{\pm 0.6}$}  &  \underline{$2.8_{\pm 0.5}$}    &  \bm{$83.2_{\pm 1.3}$}  &   \bm{$87.8_{\pm 1.5}$}    &  \underline{$0.4_{\pm 0.2}$}  &  \underline{$4.4_{\pm 0.6}$}      \\
\bottomrule
\end{tabular}}
\end{table*}

\textbf{Model Components.}
We investigate the key model components in the FairEHR-CLP, namely the CL approach and the DR layer. We maintain synthetic counterparts for data augmentation during the training phase when CL is not applied. Results from the STARR dataset, as shown in Table~\ref{tab: ablation_model_results}, reveal that removing both CL and DR results in the most significant performance degradation, averaging a 2.6\% drop in F1 and 4.1\% in AUROC across three tasks. This setup also yields the most biased predictions. The absence of either CL or DR (full w/o CL or full w/o DR) leads to only a slight decline in performance but shows a tendency towards more biased outcomes compared to those from the full model. This can be attributed to the complementary roles of CL and DR in balancing accurate predictions with fairness.

\begin{table*}[htbp]
\centering
\caption{Effects of different model components for FairEHR-CLP (full) on the STARR dataset ($\mathcal{D + L + N}$). Here, ‘w/o CL' and ‘w/o DR' represent the full model without contrastive learning and without the Dynamic Relevance layer, respectively.}
\vspace{0.1cm}
\resizebox{\linewidth}{!}{%
\label{tab: ablation_model_results}
\begin{tabular}{ccccccccccccc}
\toprule
\multirow{2}{*}{\textbf{\makecell{Model \\ Components}}} & \multicolumn{4}{c}{\textbf{Delirium}} & \multicolumn{4}{c}{\textbf{OUD}} & \multicolumn{4}{c}{\textbf{30-Day Readmission}} \\ 
\cmidrule(lr){2-5} \cmidrule(lr){6-9} \cmidrule(lr){10-13}
 & F1 ($\uparrow$) & AUROC ($\uparrow$) & EO ($\downarrow$) & EDDI ($\downarrow$) & F1 ($\uparrow$) & AUROC ($\uparrow$) & EO ($\downarrow$) & EDDI ($\downarrow$) & F1 ($\uparrow$) & AUROC ($\uparrow$) & EO ($\downarrow$) & EDDI ($\downarrow$)\\ 
\midrule
Full w/o CL + DR & $80.3_{\pm 1.4}$ & $82.1_{\pm 1.1}$ & $7.2_{\pm 0.8}$ & $5.3_{\pm 0.5}$ & $85.9_{\pm 1.7}$ & $89.7_{\pm 1.3}$ & $5.1_{\pm 0.7}$ & $4.3_{\pm 0.6}$ & $81.2_{\pm 1.2}$ & $83.5_{\pm 1.3}$ & $1.9_{\pm 0.3}$ & $5.6_{\pm 0.5}$ \\
Full w/o CL & $81.1_{\pm 1.3}$ & $83.0_{\pm 1.0}$ & $6.7_{\pm 0.7}$ & $4.5_{\pm 0.4}$ & $86.0_{\pm 1.5}$ & $89.9_{\pm 1.2}$ & $4.7_{\pm 0.6}$ & $3.8_{\pm 0.5}$ & $81.7_{\pm 1.1}$ & $84.6_{\pm 1.2}$ & $1.6_{\pm 0.2}$ & $5.1_{\pm 0.4}$ \\
Full w/o DR & $82.5_{\pm 1.2}$ & $85.4_{\pm 0.9}$ & $6.4_{\pm 0.6}$ & $4.1_{\pm 0.3}$ & $86.2_{\pm 1.3}$ & $90.3_{\pm 1.1}$ & $3.9_{\pm 0.5}$ & $3.2_{\pm 0.4}$ & $82.1_{\pm 1.0}$ & $86.2_{\pm 1.1}$ & $1.2_{\pm 0.2}$ & $4.8_{\pm 0.3}$ \\
Full  &  \bm{$84.1_{\pm 1.3}$} &  \bm{$87.3_{\pm 1.0}$} &  \underline{$5.7_{\pm 0.7}$}  &  \underline{$3.4_{\pm 0.5}$}    &  \bm{$86.3_{\pm 1.6}$}  &   \bm{$90.6_{\pm 1.4}$}    &  \underline{$3.5_{\pm 0.6}$}  &  \underline{$2.8_{\pm 0.5}$}    &  \bm{$83.2_{\pm 1.3}$}  &   \bm{$87.8_{\pm 1.5}$}    &  \underline{$0.4_{\pm 0.2}$}  &  \underline{$4.4_{\pm 0.6}$}   \\
\bottomrule
\end{tabular}}
\end{table*}

\begin{figure}[h]
\centering
\includegraphics[width=.9\textwidth]{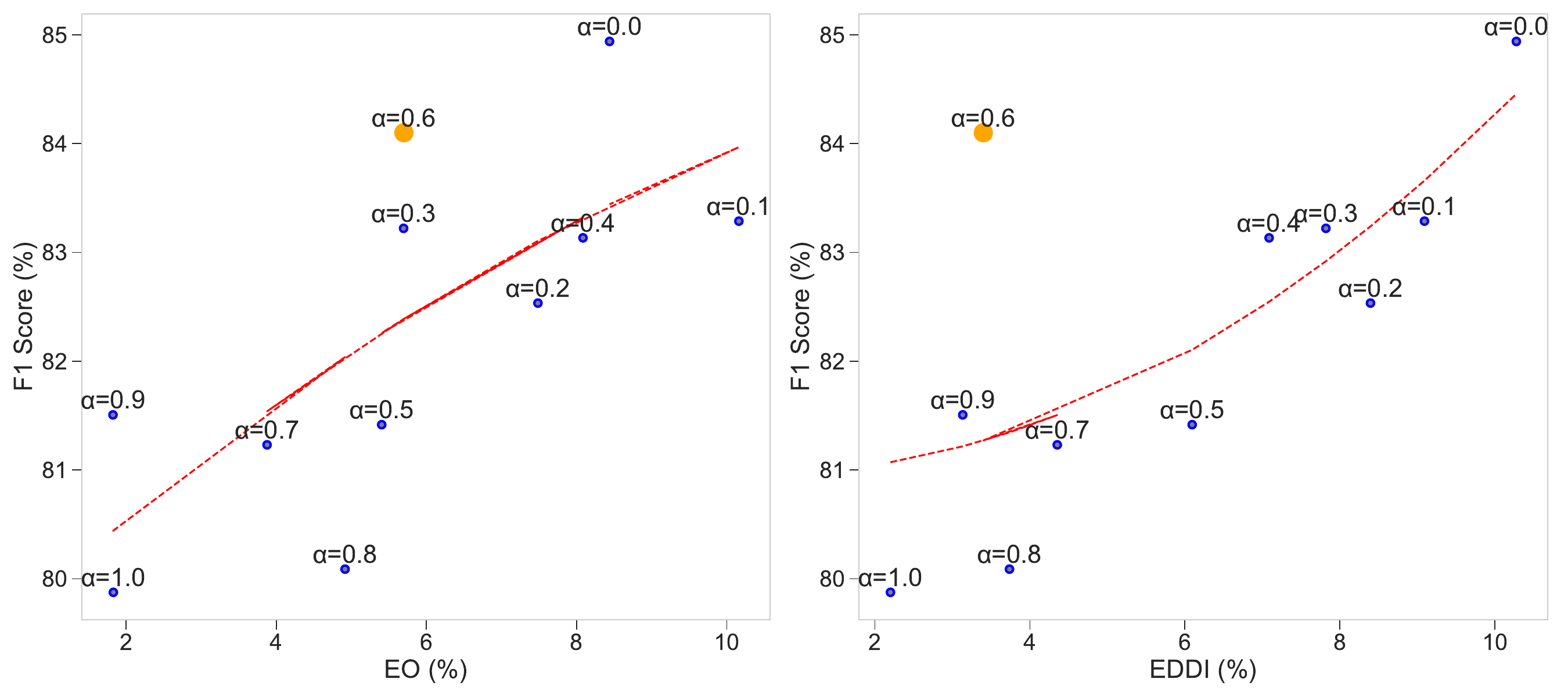}
\caption{Effect of $\alpha$ on fairness-utility trade-off in the STARR dataset (delirium task). Left: EO vs. F1; Right: EDDI vs. F1.}
\label{fig: ablation_alpha}
\end{figure}

\textbf{Effect of $\alpha$.}
We investigate the effect of $\alpha$ on the trade-off between fairness and utility across a range from 0.0 to 1.0. Figure~\ref{fig: ablation_alpha} demonstrates that, generally, a lower $\alpha$ prioritizes utility, resulting in higher F1 scores at the expense of fairness, as reflected by increased EO and EDDI values. Conversely, a higher $\alpha$ enhances fairness, evidenced by lower EO and EDDI, but leads to decreased F1 scores. The figure indicates that the optimal $\alpha=0.6$, positioned at the top-left corner of both plots, signifies an equitable compromise between fairness and utility. 

\begin{figure}[h]
\centering
\includegraphics[width=.9\textwidth]{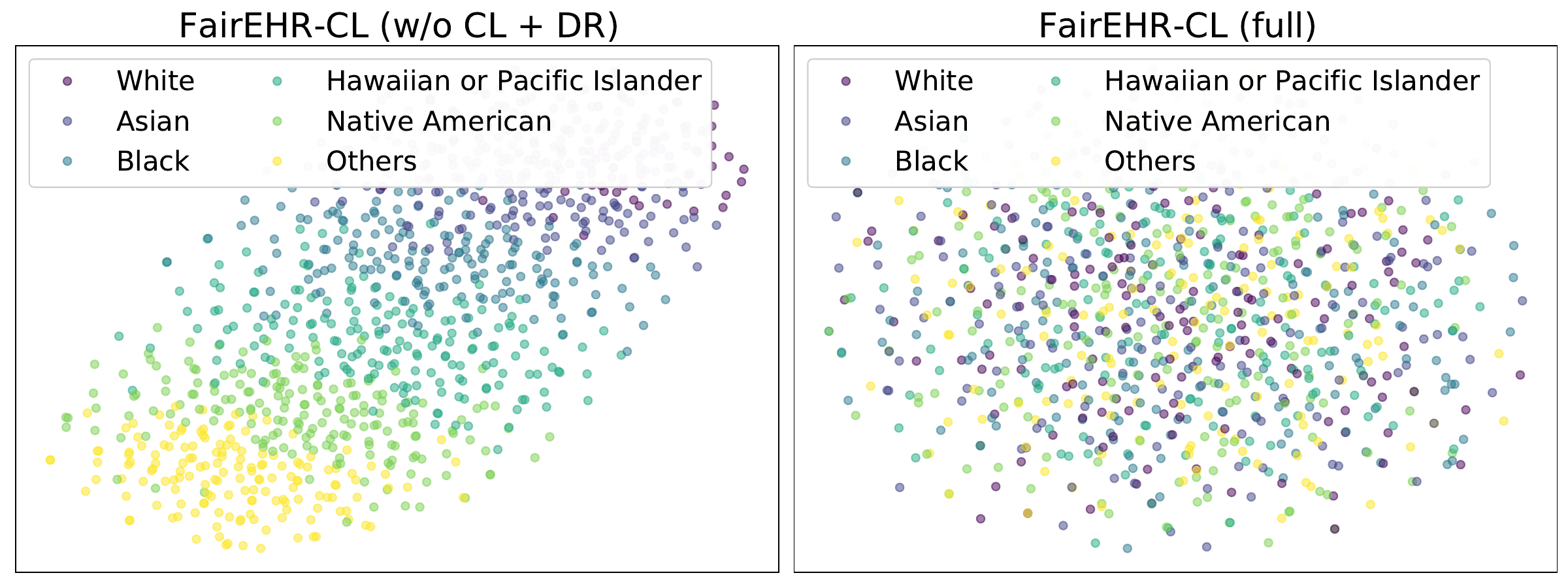}
\caption{t-SNE visualization of learned representations from FairEHR-CLP with and without bias mitigation components CL and DR on the STARR dataset w.r.t. the sensitive group race.}
\label{fig: visualization}
\end{figure}

\subsection{Visualization}
\label{tsne_visualization}
To assess the quality of the learned representations and the effectiveness of our method, we employ t-SNE~\citep{van2008visualizing} to visualize projections of 1000 patient records from the STARR test set, focusing on the sensitive attribute race, as shown in Figure~\ref{fig: visualization}. The left panel depicts a vanilla model lacking the CL and DR components, which are integral to bias mitigation in FairEHR-CLP. We observe that the vanilla model learns information about race, as the representations given by vanilla exhibit distinct clusters along racial lines. It suggests that the model may be disproportionately weighting race when forming representations. In contrast, our full FairEHR-CLP model on the right shows a more homogeneous distribution across racial groups, suggesting a reduced impact of race on the representations, thereby diminishing reliance on biased attributes and advancing towards more equitable predictions.
\vspace{-0.3cm}

\subsection{Sensitive Attributes Analysis}
\label{case_study}
We investigate the impact of our method on each sensitive attribute from a fairness perspective. Table~\ref{tab: case_individual_attributes} presents the EO and EDDI values for each sensitive attribute across three datasets. Our approach consistently demonstrates the least bias in gender, with EO as low as 1.7\% and EDDI at 1.8\%, followed by a slightly increasing bias in SES. The most biased sensitive attribute is race, exhibiting up to 5.9\% in EO and 4.7\% in EDDI. Similarly, age bias is also pronouncedly high. The variability in bias levels across different sensitive attributes and datasets underscores the impact of dataset-specific characteristics on model fairness.

\begin{table}[htbp]
\centering
\caption{Fairness evaluation of FairEHR-CLP across individual sensitive attributes in three datasets, averaged over three tasks. Bold values represent the least bias, while underlined values indicate the most bias among sensitive attributes.}
\label{tab: case_individual_attributes}
\vspace{0.1cm}

\begin{tabular}{ccccccc}
\toprule
\multirow{2}{*}{\textbf{Attributes}} & \multicolumn{2}{c}{\textbf{STARR}} & \multicolumn{2}{c}{\textbf{MIMIC-III}} & \multicolumn{2}{c}{\textbf{MIMIC-IV}} \\ 
\cmidrule(lr){2-3} \cmidrule(lr){4-5} \cmidrule(lr){6-7}
  & EO ($\downarrow$) & EDDI ($\downarrow$) & EO ($\downarrow$) & EDDI ($\downarrow$) &  EO ($\downarrow$) & EDDI ($\downarrow$)\\ 
\midrule
Gender    & \bm{$1.7_{\pm 0.5}$} & \bm{$2.4_{\pm 0.5}$} & \bm{$3.3_{\pm 0.3}$} & \bm{$1.8_{\pm 0.3}$} & \bm{$2.8_{\pm 0.4}$} & \bm{$3.3_{\pm 0.4}$} \\
Race      & \underline{$5.2_{\pm 0.8}$} & \underline{$4.7_{\pm 0.7}$} & \underline{$5.9_{\pm 0.6}$} & \underline{$3.2_{\pm 0.4}$} & $3.8_{\pm 0.6}$ & $4.2_{\pm 0.7}$ \\
Ethnicity & $3.0_{\pm 0.5}$ & $3.6_{\pm 0.3}$ & $4.4_{\pm 0.4}$ & $2.6_{\pm 0.6}$ & $3.5_{\pm 0.3}$ & $3.9_{\pm 0.6}$ \\
Age       & $3.5_{\pm 0.3}$ & $3.7_{\pm 0.4}$ & $4.6_{\pm 0.4}$ & $3.0_{\pm 0.7}$ & \underline{$4.1_{\pm 0.8}$} & \underline{$4.4_{\pm 0.8}$} \\
SES       & $2.6_{\pm 0.4}$ & $3.1_{\pm 0.6}$ & $3.8_{\pm 0.3}$ & $2.4_{\pm 0.5}$ & $3.3_{\pm 0.4}$ & $3.7_{\pm 0.5}$ \\
\bottomrule
\end{tabular}
\end{table}

\section{Discussion}
In this paper, we have presented a novel approach to address the challenges of fairness in clinical predictions using EHRs. Our findings suggest that the FairEHR-CLP framework, which integrates patient demographics, longitudinal data, and clinical notes through a unique two-stage process: synthetic counterpart generation and fairness-aware predictions with CL, significantly reduces disparities in error rates across different demographic subgroups. This improvement is critical in the context of healthcare, where equitable treatment and diagnosis are paramount. The integration of contrastive learning in fairness-aware predictions, combined with our novel fairness metric, represents a substantial advancement in the pursuit of equitable healthcare outcomes. 

\paragraph{Limitations and Future Work.}
A concern in our study is the quality of synthetic data generated. Inaccuracies in capturing the complexity of real patient data could limit the model's effectiveness in mitigating biases. Future research should explore diverse synthetic data generation techniques, especially for longitudinal data and notes, to identify those that most accurately mirror the statistical characteristics of real data. Additionally, our approach encounters challenges with ambiguous categories in sensitive attributes, such as ‘Unknown’ or ‘Other’. Refining categorization strategies is crucial to address biases more precisely. We will also extend our experiments to various clinical contexts, thereby enhancing the robustness and adaptability of our approach.

\section{Broader Impacts}
This paper introduces a general framework aimed at enhancing fairness in clinical predictions using multimodal EHRs by addressing social biases from demographic factors. Our approach highlights the potential for more equitable healthcare outcomes through ethically conscious AI, underscoring the importance of responsible usage. FairEHR-CLP offers a promising avenue to close the disparities gap in health outcomes by ensuring more accurate and unbiased healthcare predictive models, paving the way for a more inclusive future in medical decision-making.

\section*{Acknowledgments}
This project was supported by grant number R01HS024096 from the Agency for Healthcare Research and Quality. The content is solely the responsibility of the authors and does not necessarily represent the official views of the Agency for Healthcare Research and Quality.


\bibliography{sample}

\appendix
\newpage
\section{EHR Data Examples}
\label{synthetic_example}
We provide a sample of EHR data from MIMIC-IV for one patient, including both real and synthetic data, encompassing static demographic features, longitudinal data, and clinical notes.

\begin{figure}[ht]
    \centering
    \subfigure[Real demographic features.]{
        \includegraphics[width=0.4\textwidth]{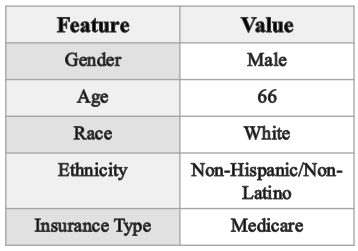}
    }
    \hfill
    \hspace{5pt}
    \subfigure[Synthetic demographic features.]{
        \includegraphics[width=0.41\textwidth]{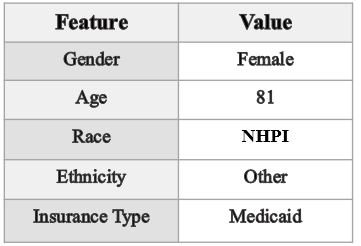}
    }
    \caption{Demographic examples (real and synthetic) from an EHR data sample. NHPI denotes Native Hawaiians and Pacific Islanders.}
    \label{fig: demographic_example}
\end{figure}

\noindent \textbf{Demographics.} Figure~\ref{fig: demographic_example} provides an example of real and synthetic demographic features for a patient. 

\begin{figure}[ht]
    \centering
    \subfigure[Real longitudinal features.]{
        \includegraphics[width=0.65\textwidth]{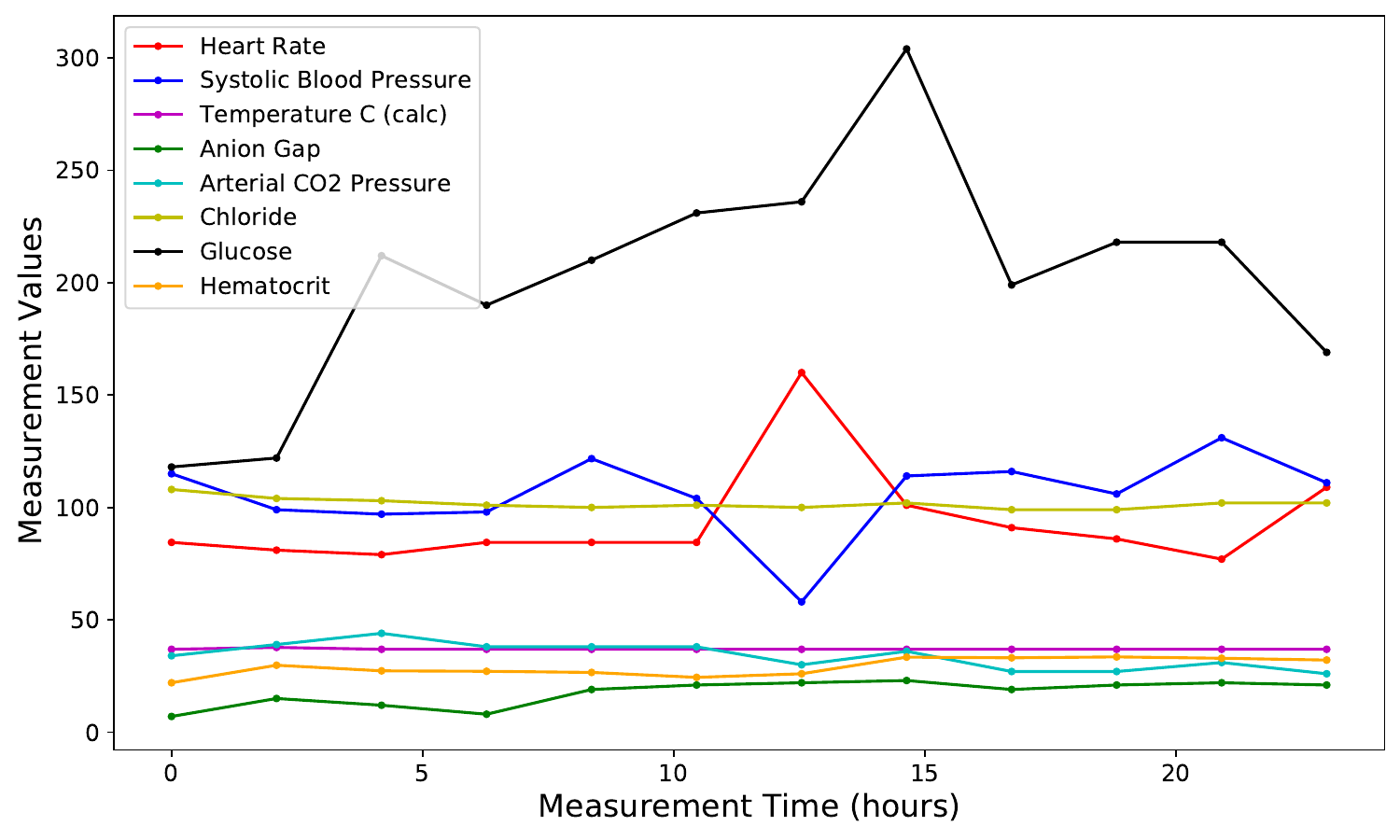}
    }
    \hfill
    \hspace{5pt}
    \subfigure[Synthetic longitudinal features.]{
        \includegraphics[width=0.65\textwidth]{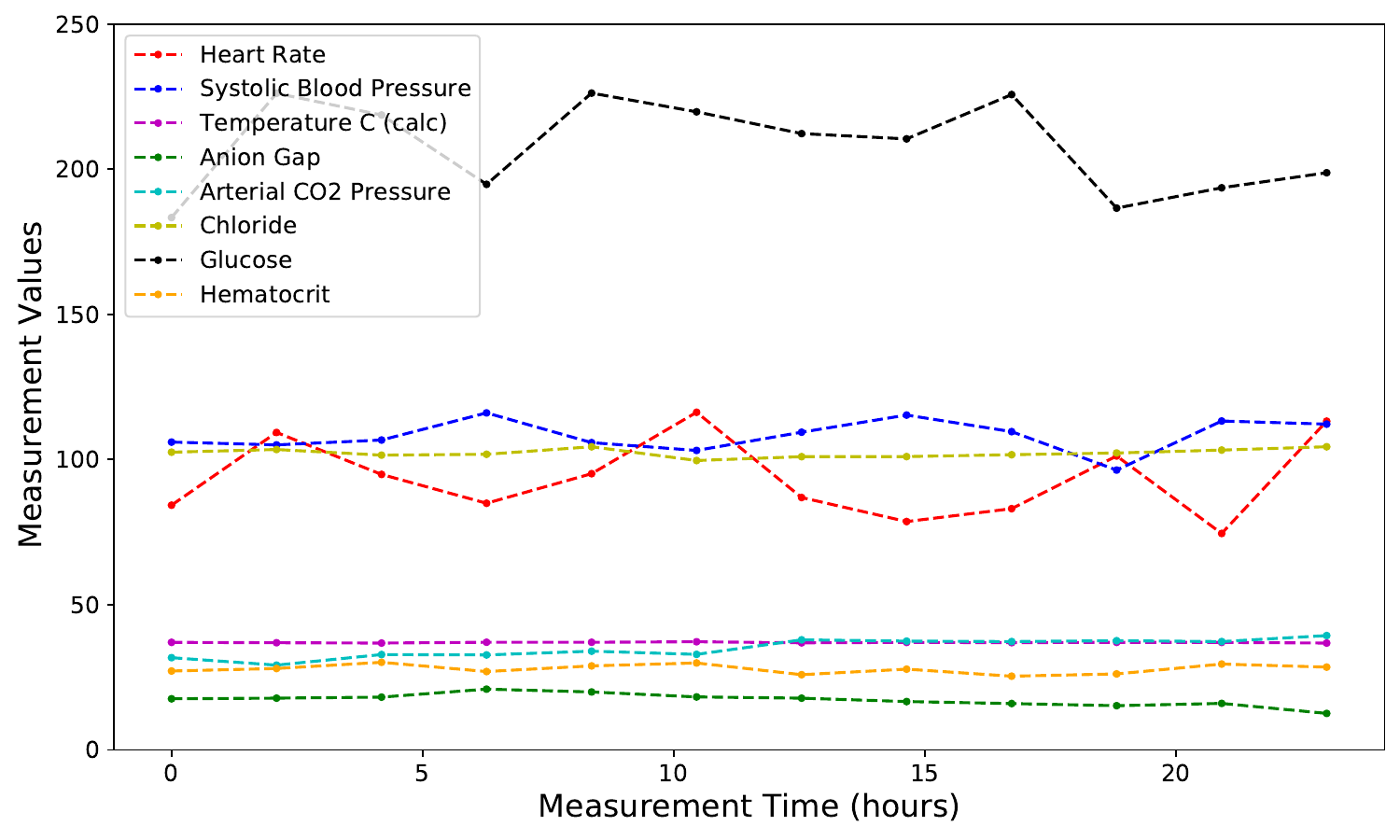}
    }
    \caption{Longitudinal feature examples (real and synthetic) from an EHR data sample.}
    \label{fig: longitudinal_example}
\end{figure}

\noindent \textbf{Longitudinal Data.} Figure~\ref{fig: longitudinal_example} presents an example of real and synthetic longitudinal data for a patient.

\noindent \textbf{Notes.} We provide the following examples of real and synthetic clinical notes for the patient described earlier. The texts in bold indicate the patient's primary medical or health conditions.

Real: The patient exhibited a \textbf{progressive exacerbation of dyspnea and edema} over four days, ultimately found in a tripod position with a resting arterial oxygen saturation of 90\%. Initially managed as a COPD exacerbation and later excluding non-ST-elevation myocardial infarction (NSTE-ACS), the patient was stabilized in the ICU with \textbf{BiPAP support}. Subsequent cardiac catheterization identified \textbf{multivessel coronary artery disease}, including \textbf{in-stent stenosis} in the left anterior descending artery. Despite these complications, the patient remained hemodynamically stable in \textbf{normal sinus rhythm} and was subsequently shifted for \textbf{revascularization evaluation}.

Synthetic: The patient arrived with an \textbf{escalating severity in breathing difficulty and swelling} over four days, observed in a respiratory distress posture with an oxygen saturation level at 90\%. Initially treated for a chronic obstructive pulmonary disease flare-up, myocardial infarction without ST-elevation was later ruled out. The patient was maintained in a stable condition under \textbf{BiPAP respiratory support} in the intensive care unit. Cardiac catheterization conducted recently revealed a \textbf{complex coronary artery disease}, notably including a \textbf{narrowed segment within a stent} in the left anterior descending artery. Notwithstanding these heart-related complexities, the patient's hemodynamic status was stable with a \textbf{normal heart rhythm}, leading to a transfer for further assessment and planning for \textbf{revascularization therapy}.

\section{Synthetic Notes Review Guidelines}
\label{notes_guideline}
For quality assurance, we randomly select 100 synthetic patient notes from each of the three datasets. The manual review process adheres to the following principles:
\begin{enumerate}
\item[(1)] \textbf{Exclusion of Demographic Factors}: Demographic identifiers such as gender, race, age, ethnicity, and socioeconomic status (SES) associated with insurance type are excluded to ensure the notes primarily focus on health conditions, aligning with our objective to mitigate social bias stemming from demographic factors in clinical predictions.
\item[(2)] \textbf{Inclusion of Major Treatments and Diagnoses}: We verify the presence and accuracy of essential health information, including diagnoses, treatments, and medical history, to ensure the synthetic notes retain critical medical content for predictive modeling relevance.
\item[(3)] \textbf{Consistency with Real Records:} The synthetic notes are compared against authentic clinical records to ascertain their fidelity in mirroring the structure, terminology, and clinical reasoning typical of real medical documentation.
\end{enumerate}

\section{Notations}
\label{notations}
All the notations corresponding to the FairEHR-CLP framework are summarized in Table~\ref{tab: notations1} and Table~\ref{tab: notations2}.

\begin{table}[ht]
\centering
\caption{Notation definitions in FairEHR-CLP (Part 1).}
\vspace{0.2cm}
 \resizebox{\linewidth}{!}{%
\begin{tabular}{c c l}
\toprule
\textbf{Reference} & \textbf{Notation} & \quad \quad \quad \quad \quad \quad \quad \quad \quad \quad \quad \textbf{Description} \\
\midrule
\multirow{6}{*}{\textbf{\makecell{Section 3.1 \\ Problem \\Formulation}}} & $\mathscr{D}$ & Dataset with patient data, labels, and sensitive attributes \\
& $x_k \in \mathcal{X}$ & Input features from demographics, longitudinal data, and clinical notes \\
& $y_k \in \{0, 1\} \subseteq \mathcal{Y}$ & Binary target label for patient outcomes \\
& $s_k \in \mathcal{S}$ & Sensitive attributes from demographic features \\
& $\mathcal{S}$ & Set of sensitive attributes including gender, race, ethnicity, age, and SES \\
& $f: \mathcal{X} \rightarrow \mathcal{Y}$ & Prediction model from features to outcomes \\
\midrule
\multirow{16}{*}{\textbf{\makecell{Section 3.2 \\ Longitudinal Data \\ EHR-GAN}}} & $G$ & Generator\\
& $G_{e}$ & Encoder component of generator \\
& $G_{d}$ & Decoder component of generator \\
& $D$ & Discriminator in EHR-GAN \\
& $x$ & Input data to encoder \\
& $z$ & Latent space representation from encoder \\
& $v$ & Random noise input to decoder \\
& $\hat{x}$ & Synthetic data generated by decoder \\
& $l_{dis}$ & Discriminative loss by discriminator \\
& $l_{adv}$ & Adversarial loss for generator \\
& $l_{fm}$ & Feature matching loss for generator \\
& $\beta_0, \beta_1, \beta_2$ & Weighting coefficients for loss components \\
& $p_z(z)$ & Prior distribution over latent space \\
& $p_{x}(x)$ & Distribution of real data \\
& $f(\cdot)$ & Output of intermediate layer in discriminator \\
& $y_i$ & Label indicating real or synthetic data \\
& & \quad \quad \quad \quad \quad \quad \quad \quad \quad \quad \quad \quad \quad \quad \quad \quad \quad \quad \quad \quad Continued on next page \\
\bottomrule
\end{tabular}}
\label{tab: notations1}
\end{table}

\begin{table}[ht]
\centering
\caption{Notation definitions in FairEHR-CLP (Part 2).}
\vspace{0.2cm}
 \resizebox{\linewidth}{!}{%
\begin{tabular}{c c l}
\toprule
\textbf{Reference} & \textbf{Notation} & \quad \quad \quad \quad \quad \quad \quad \quad \quad \quad \quad \textbf{Description} \\
\midrule
\multirow{21}{*}{\textbf{\makecell{Section 3.3 \\ Fairness-aware \\ Prediction with \\ Contrastive Learning}}} & $x^{+}$ & Positive samples: synthetic counterparts \\
& $x^{-}$ & Negative samples: other patient data in minibatch \\
& $e_d$ & Encoded demographic data \\
& $e_l$ & Encoded longitudinal data \\
& $e_n$ & Encoded clinical notes \\
& $F_{\text{fusion}}$ & MLP-based fusion function \\
& $\theta_{\text{fusion}}$ & Trainable parameters in fusion layer \\
& $e$, $e^{syn}$ & Integrated representations for real and synthetic data \\
& $F_{\text{DR}}$ & Dynamic Relevance (DR) layer function \\
& $w$ & Adjustable weights in DR layer \\
& $\sigma$ & Sigmoid function \\
& $e^{adj}$, $e^{adj, syn}$ & Adjusted embeddings post-DR layer \\
& $l_{CF}$ & Fairness-oriented contrastive loss \\
& $l_{CE}$ & Cross entropy loss \\
& $N$ & Number of embeddings in minibatch \\
& $\tau$ & Temperature parameter \\
& $\gamma$ & Regularization parameter \\
& $\mu_{\text{syn}}^{adj}$ & Mean of adjusted synthetic embeddings \\
& $y_k$ & True label for each real embedding \\
& $C(e^{adj}_{k})$ & softmax probability of predicted class \\
& $\alpha$ & Parameter balancing fairness and performance \\
\bottomrule
\end{tabular}}
\label{tab: notations2}
\end{table}

\section{Implementation Details}
\label{app_implementation}
All of the experiments are conducted on four NVIDIA A100 GPUs. We apply a random train/test split in an 80\%/20\% ratio for each dataset. In training our EHR-GAN, we primarily adhere to the experimental settings of the baseline EHR-M-GAN as described in \citet{li2023generating}, omitting the discrete-valued time-series data and focusing solely on continuous longitudinal data. Maximum Mean Discrepancy (MMD) is employed to assess the similarity between real and synthetic data, aiding in the adjustment of hyperparameters in EHR-GAN for quality control. For detailed implementation specifics, please refer to \citet{li2023generating}. Based on the results in the original paper and our experiments, we set the MMD threshold at 0.68 to ensure a reasonable quality of synthetic longitudinal data. After the first stage of FairEHR-CLP, which involves synthetic counterpart generation, and considering that we have both synthetic and real data for each patient in the training set (demographics, longitudinal, and notes), we employ fairness-aware predictions with CL. The Adam optimizer is utilized with its default parameters for optimization. The hyperparameter search space for all datasets is detailed in Table~\ref{tab: hp_space}. Hyperparameter optimization is conducted via random search.
\vspace{-0.5cm}
\begin{table}[ht]
  \centering
  \caption{Hyperparameter search space of FairEHR-CLP on three datasets.}
  \vspace{0.2cm}
  \label{tab: hp_space}
  \begin{tabular}{cc}
    \toprule
    Hyperparameters & Search Space\\
    \midrule
     Batch size & [16, 32, 64, 128, 256] \\
     Learning rate & [$1\mathrm{e}{-5}, 5\mathrm{e}{-5}$, $1\mathrm{e}{-6}$, $5\mathrm{e}{-6}$] \\
     \# of epochs &  [20, 30, 50] \\
     $\tau$ & [0.1, 0.3, 0.5, 0.7] \\
     $\lambda$ & [0.3, 0.4, 0.5, 0.6, 0.7] \\
  \bottomrule
\end{tabular}
\end{table}

\section{Datasets}
\label{app_datasets}
We summarize the clinical predictors, including vital signs and laboratory measurements, used in the MIMIC-III/IV and STARR datasets, in Table~\ref{tab: clinical_predictors_mimic} and Table~\ref{tab: clinical_predictors_starr}, respectively. These predictors are used for all three prediction tasks: classifying delirium, OUD, and 30-day readmission. Due to the absence of explicit codes for identifying surgical patients in the MIMIC-III/IV datasets, we extract patient data from the Surgical Intensive Care Unit (SICU). Delirium refers to a condition characterized by confusion and a reduced ability to maintain attention and clear awareness, with its incidence increasing with age~\citep{wilson2020delirium}. Bias could arise from healthcare professionals’ age-related stereotypes, leading to underdiagnosis in older patients or overdiagnosis in those with pre-existing cognitive impairments, which could affect treatment decisions and ultimately patient recovery. OUD is a medical condition characterized by the problematic use of opioid medications, commonly prescribed for pain relief, and can lead to a high risk of dependence and misuse. OUD can be influenced by biases related to prescribing practices, such as biases based on patients' race or socioeconomic status, which might affect the likelihood of being prescribed opioids, the dosage, or the duration of use, potentially leading to disparities in the risk of developing OUD. Lastly, 30-day readmission is defined as the rehospitalization of a patient within 30 days following their discharge from a hospital, serving as an important indicator of the quality of care and patient outcomes. For example, elderly patients might receive less comprehensive discharge planning or follow-up care due to assumptions about their support systems or ability to manage their own care, leading to higher readmission rates.

\begin{table}[ht]
\centering
\caption{Summary of clinical predictors in longitudinal data for MIMIC-III/IV datasets.}
\vspace{0.2cm}
\label{tab: clinical_predictors_mimic}
 \resizebox{\linewidth}{!}{%
\begin{tabular}{lc}
\hline
\textbf{Category} & \textbf{Predictors} \\ \hline
Vital Signs & \makecell{Heart Rate, Systolic Blood Pressure, Diastolic Blood Pressure, Mean Blood Pressure, \\ Respiratory Rate, Body Temperature, Oxygen Saturation} \\ \hline
Blood Gases & \makecell{Arterial Base Excess, Arterial Carbon Dioxide Pressure, \\ Arterial Oxygen Pressure, Arterial pH} \\ \hline
Renal Function & Blood Urea Nitrogen, Creatinine \\ \hline
Metabolic Panel & \makecell{Ionized Calcium, Serum Chloride, Serum Glucose, Fingerstick Glucose, Anion Gap,\\ Serum Bicarbonate, Magnesium, Phosphorus, Serum Potassium, Serum Sodium} \\ \hline
Hematology & \makecell{Serum Hematocrit, Hemoglobin, \\ Platelet Count, White Blood Cell Count} \\ \hline
\end{tabular}}
\end{table}

\begin{table}[ht]
\centering
\caption{Summary of clinical predictors in longitudinal data for the STARR dataset.}
\vspace{0.2cm}
\label{tab: clinical_predictors_starr}
 \resizebox{\linewidth}{!}{%
\begin{tabular}{lc}
\hline
\textbf{Category} & \textbf{Predictors} \\ \hline
Vital Signs & \makecell{Heart Rate, Pulse, Respiratory Rate, Oxygen Saturation, \\ Body Temperature, Systolic Blood Pressure, Diastolic Blood Pressure} \\ \hline
Blood Gases & \makecell{CO2, Anion Gap} \\ \hline
Renal Function & \makecell{Blood Urea Nitrogen, Creatinine} \\ \hline
Metabolic Panel & \makecell{Calcium, Chloride, Glucose, Potassium, Sodium} \\ \hline
Hematology & \makecell{Hematocrit, Hemoglobin, Mean Corpuscular Volume, \\ Mean Corpuscular Hemoglobin, White Blood Cell Count, \\ Platelet Count, Red Blood Cell Count, Red Cell Distribution Width, \\ Mean Corpuscular Hemoglobin Concentration} \\ \hline
Liver Function & \makecell{ALT (SGPT), Albumin} \\ \hline
\end{tabular}}
\end{table}

\newpage

\section{Ablation Study}
\label{app_ablation}
\textbf{Data Modalities.} Table~\ref{tab: ablation_data_results_mimiciii} and Table~\ref{tab: ablation_data_results_mimiciv} demonstrate the impact of different data modalities on the performance of our FairEHR-CLP method for the MIMIC-III and MIMIC-IV datasets, respectively. Similar to the trends observed in the STARR dataset, combining demographic ($\mathcal{D}$) and longitudinal ($\mathcal{L}$) data surpasses the mix of $\mathcal{D}$ with clinical notes. In MIMIC-III, using the complete dataset ($\mathcal{D + L + N}$) results in a 5.8\% increase in F1 and a 4.9\% improvement in AUROC compared to the second-best combination ($\mathcal{D + L}$). Likewise, for MIMIC-IV, employing the full dataset ($\mathcal{D + L + N}$) leads to a 2.0\% enhancement in F1 and a 2.4\% increase in AUROC over the second-best results ($\mathcal{D + L}$). In terms of fairness metrics, the full dataset consistently yields lower EO and EDDI values compared to the use of partial data. This highlights the effectiveness of comprehensive patient representation in achieving more equitable predictions.

\begin{table*}[htbp]
\centering
\caption{Effects of different data modalities as inputs for FairEHR-CLP on the MIMIC-III dataset.}
\vspace{0.2cm}
\resizebox{\linewidth}{!}{%
\label{tab: ablation_data_results_mimiciii}
\begin{tabular}{ccccccccccccc}
\toprule
\multirow{2}{*}{\textbf{\makecell{Data \\ Modalities}}} & \multicolumn{4}{c}{\textbf{Delirium}} & \multicolumn{4}{c}{\textbf{OUD}} & \multicolumn{4}{c}{\textbf{30-Day Readmission}} \\ 
\cmidrule(lr){2-5} \cmidrule(lr){6-9} \cmidrule(lr){10-13}
 & F1 ($\uparrow$) & AUROC ($\uparrow$) & EO ($\downarrow$) & EDDI ($\downarrow$) & F1 ($\uparrow$) & AUROC ($\uparrow$) & EO ($\downarrow$) & EDDI ($\downarrow$) & F1 ($\uparrow$) & AUROC ($\uparrow$) & EO ($\downarrow$) & EDDI ($\downarrow$)\\ 
\midrule
$\mathcal{D}$ &  $77.3_{\pm 1.8}$ & $79.6_{\pm 1.7}$ & $8.1_{\pm 1.0}$ & $5.8_{\pm 0.7}$ & $80.5_{\pm 1.6}$ & $84.3_{\pm 1.4}$ & $5.0_{\pm 0.9}$ & $4.4_{\pm 0.8}$ & $80.6_{\pm 1.5}$ & $83.4_{\pm 1.3}$ & $6.1_{\pm 0.6}$ & $6.3_{\pm 0.7}$ \\

$\mathcal{D + L}$ & $81.0_{\pm 1.5}$ & $85.3_{\pm 1.3}$ & $6.5_{\pm 0.8}$ & $4.5_{\pm 0.6}$ & $83.8_{\pm 1.7}$ & $87.9_{\pm 1.5}$ & $4.2_{\pm 0.7}$ & $3.3_{\pm 0.6}$ & $83.9_{\pm 1.4}$ & $87.1_{\pm 1.2}$ & $5.3_{\pm 0.5}$ & $5.0_{\pm 0.6}$ \\

$\mathcal{D + N}$ &  $79.8_{\pm 1.6}$ & $83.3_{\pm 1.4}$ & $7.0_{\pm 0.7}$ & $5.1_{\pm 0.6}$ & $82.4_{\pm 1.8}$ & $86.5_{\pm 1.6}$ & $4.6_{\pm 0.6}$ & $3.8_{\pm 0.7}$ & $80.7_{\pm 1.3}$ & $84.5_{\pm 1.1}$ & $5.7_{\pm 0.4}$ & $5.1_{\pm 0.8}$ \\

$\mathcal{D + L + N}$  & \bm{$85.5_{\pm 1.2}$} &  \bm{$89.7_{\pm 1.1}$} & \underline{$6.2_{\pm 0.3}$} & \underline{$3.8_{\pm 0.5}$} & \bm{$89.4_{\pm 1.4}$} & \bm{$91.9_{\pm 1.5}$} & \underline{$3.7_{\pm 0.5}$} & \underline{$2.0_{\pm 0.4}$} & \bm{$88.2_{\pm 1.3}$} & \bm{$91.4_{\pm 1.1}$} & \underline{$3.3_{\pm 0.4}$} & \underline{$2.1_{\pm 0.6}$}  \\
\bottomrule
\end{tabular}}
\end{table*}

\begin{table*}[htbp]
\centering
\caption{Effects of different data modalities as inputs for FairEHR-CLP on the MIMIC-IV dataset.}
\vspace{0.2cm}
\resizebox{\linewidth}{!}{%
\label{tab: ablation_data_results_mimiciv}
\begin{tabular}{ccccccccccccc}
\toprule
\multirow{2}{*}{\textbf{\makecell{Data \\ Modalities}}} & \multicolumn{4}{c}{\textbf{Delirium}} & \multicolumn{4}{c}{\textbf{OUD}} & \multicolumn{4}{c}{\textbf{30-Day Readmission}} \\ 
\cmidrule(lr){2-5} \cmidrule(lr){6-9} \cmidrule(lr){10-13}
 & F1 ($\uparrow$) & AUROC ($\uparrow$) & EO ($\downarrow$) & EDDI ($\downarrow$) & F1 ($\uparrow$) & AUROC ($\uparrow$) & EO ($\downarrow$) & EDDI ($\downarrow$) & F1 ($\uparrow$) & AUROC ($\uparrow$) & EO ($\downarrow$) & EDDI ($\downarrow$)\\ 
\midrule
$\mathcal{D}$ & $75.6_{\pm 1.8}$ & $77.8_{\pm 1.7}$ & $8.5_{\pm 1.0}$ & $6.2_{\pm 0.7}$ & $81.9_{\pm 1.6}$ & $84.9_{\pm 1.4}$ & $5.4_{\pm 0.8}$ & $4.7_{\pm 0.6}$ & $76.3_{\pm 1.7}$ & $79.5_{\pm 1.5}$ & $6.7_{\pm 0.7}$ & $6.9_{\pm 0.8}$ \\

$\mathcal{D + L}$ & $78.2_{\pm 1.5}$ & $81.6_{\pm 1.3}$ & $7.0_{\pm 0.8}$ & $5.1_{\pm 0.6}$ & $83.7_{\pm 1.7}$ & $87.8_{\pm 1.6}$ & $4.7_{\pm 0.7}$ & $3.6_{\pm 0.5}$ & $78.5_{\pm 1.6}$ & $82.4_{\pm 1.4}$ & $6.0_{\pm 0.6}$ & $5.7_{\pm 0.7}$ \\

$\mathcal{D + N}$ & $76.9_{\pm 1.6}$ & $80.3_{\pm 1.5}$ & $7.8_{\pm 0.9}$ & $5.6_{\pm 0.7}$ & $82.5_{\pm 1.8}$ & $86.7_{\pm 1.5}$ & $5.1_{\pm 0.6}$ & $4.2_{\pm 0.6}$ & $77.1_{\pm 1.5}$ & $80.7_{\pm 1.3}$ & $6.4_{\pm 0.8}$ & $6.1_{\pm 0.9}$ \\

$\mathcal{D + L + N}$  &  \bm{$78.8_{\pm 1.2}$} &  \bm{$82.4_{\pm 1.0}$} & \underline{$6.1_{\pm 0.4}$} & \underline{$3.5_{\pm 0.3}$} & \bm{$84.8_{\pm 1.6}$} & \bm{$88.9_{\pm 1.5}$} & \underline{$1.5_{\pm 0.3}$} & \underline{$3.0_{\pm 0.6}$} & \bm{$81.6_{\pm 1.8}$} & \bm{$86.4_{\pm 1.6}$} & \underline{$2.8_{\pm 0.7}$} & \underline{$5.2_{\pm 0.9}$}  \\
\bottomrule
\end{tabular}}
\end{table*}

\newpage 
\noindent \textbf{Model Components.} Table~\ref{tab: ablation_model_results_mimiciii} and Table~\ref{tab: ablation_model_results_mimiciv} demonstrate the impact of different model components on FairEHR when employing the full dataset for the MIMIC-III and MIMIC-IV datasets, respectively. During the training phase, synthetic counterparts are maintained for data augmentation when CL is not applied. For both datasets, the removal of both CL and DR leads to the most significant performance decline. Specifically, for MIMIC-III, the configuration without CL and DR (Full w/o CL + DR) results in a performance decrease of 2.7\% in F1 and 3.3\% in AUROC. For MIMIC-IV, the same configuration leads to a decrease of 4.2\% in F1 and 4.4\% in AUROC. In this case, it yields the most biased predictions with higher EO and EDDI values, while removing CL or DR moderately reduces performance but slightly increases fairness metrics. 

\begin{table*}[htbp]
\centering
\caption{Effects of different model components for FairEHR-CLP (full) on the MIMIC-III dataset ($\mathcal{D + L + N}$).}
\vspace{0.2cm}
\resizebox{\linewidth}{!}{%
\label{tab: ablation_model_results_mimiciii}
\begin{tabular}{ccccccccccccc}
\toprule
\multirow{2}{*}{\textbf{\makecell{Model \\ Components}}} & \multicolumn{4}{c}{\textbf{Delirium}} & \multicolumn{4}{c}{\textbf{OUD}} & \multicolumn{4}{c}{\textbf{30-Day Readmission}} \\ 
\cmidrule(lr){2-5} \cmidrule(lr){6-9} \cmidrule(lr){10-13}
 & F1 ($\uparrow$) & AUROC ($\uparrow$) & EO ($\downarrow$) & EDDI ($\downarrow$) & F1 ($\uparrow$) & AUROC ($\uparrow$) & EO ($\downarrow$) & EDDI ($\downarrow$) & F1 ($\uparrow$) & AUROC ($\uparrow$) & EO ($\downarrow$) & EDDI ($\downarrow$)\\ 
\midrule
Full w/o CL + DR & $83.2_{\pm 1.3}$ & $86.4_{\pm 1.2}$ & $7.0_{\pm 0.8}$ & $5.1_{\pm 0.6}$ & $87.1_{\pm 1.6}$ & $89.4_{\pm 1.4}$ & $4.4_{\pm 0.7}$ & $3.9_{\pm 0.5}$ & $85.8_{\pm 1.2}$ & $88.6_{\pm 1.3}$ & $4.8_{\pm 0.5}$ & $3.9_{\pm 0.6}$ \\

Full w/o CL & $83.4_{\pm 1.1}$ & $86.9_{\pm 1.0}$ & $6.6_{\pm 0.7}$ & $4.7_{\pm 0.4}$ & $88.3_{\pm 1.5}$ & $90.6_{\pm 1.3}$ & $4.1_{\pm 0.6}$ & $3.4_{\pm 0.4}$ & $86.9_{\pm 1.1}$ & $89.8_{\pm 1.2}$ & $4.3_{\pm 0.4}$ & $3.6_{\pm 0.5}$ \\

Full w/o DR & $84.2_{\pm 1.0}$ & $87.5_{\pm 0.9}$ & $6.3_{\pm 0.6}$ & $4.3_{\pm 0.3}$ & $88.9_{\pm 1.6}$ & $91.1_{\pm 1.2}$ & $3.9_{\pm 0.5}$ & $3.1_{\pm 0.3}$ & $87.5_{\pm 1.0}$ & $90.2_{\pm 1.3}$ & $3.4_{\pm 0.3}$ & $2.9_{\pm 0.4}$ \\

Full  &  \bm{$85.5_{\pm 1.2}$} &  \bm{$89.7_{\pm 1.1}$} & \underline{$6.2_{\pm 0.3}$} & \underline{$3.8_{\pm 0.5}$} & \bm{$89.4_{\pm 1.4}$} & \bm{$91.9_{\pm 1.5}$} & \underline{$3.7_{\pm 0.5}$} & \underline{$2.0_{\pm 0.4}$} & \bm{$88.2_{\pm 1.3}$} & \bm{$91.4_{\pm 1.1}$} & \underline{$3.3_{\pm 0.4}$} & \underline{$2.1_{\pm 0.6}$}  \\
\bottomrule
\end{tabular}}
\end{table*}

\begin{table*}[htbp]
\centering
\caption{Effects of different model components for FairEHR-CLP (full) on the MIMIC-IV dataset ($\mathcal{D + L + N}$).}
\vspace{0.2cm}
\resizebox{\linewidth}{!}{%
\label{tab: ablation_model_results_mimiciv}
\begin{tabular}{ccccccccccccc}
\toprule
\multirow{2}{*}{\textbf{\makecell{Model \\ Components}}} & \multicolumn{4}{c}{\textbf{Delirium}} & \multicolumn{4}{c}{\textbf{OUD}} & \multicolumn{4}{c}{\textbf{30-Day Readmission}} \\ 
\cmidrule(lr){2-5} \cmidrule(lr){6-9} \cmidrule(lr){10-13}
 & F1 ($\uparrow$) & AUROC ($\uparrow$) & EO ($\downarrow$) & EDDI ($\downarrow$) & F1 ($\uparrow$) & AUROC ($\uparrow$) & EO ($\downarrow$) & EDDI ($\downarrow$) & F1 ($\uparrow$) & AUROC ($\uparrow$) & EO ($\downarrow$) & EDDI ($\downarrow$)\\ 
\midrule
Full w/o CL + DR & $76.4_{\pm 1.3}$ & $79.8_{\pm 1.2}$ & $7.5_{\pm 0.9}$ & $6.3_{\pm 0.7}$ & $80.5_{\pm 1.7}$ & $84.9_{\pm 1.6}$ & $5.2_{\pm 0.8}$ & $4.5_{\pm 0.6}$ & $78.3_{\pm 1.7}$ & $82.2_{\pm 1.5}$ & $3.5_{\pm 0.6}$ & $6.6_{\pm 0.8}$ \\

Full w/o CL & $76.8_{\pm 1.1}$ & $80.9_{\pm 1.1}$ & $6.9_{\pm 0.8}$ & $5.7_{\pm 0.5}$ & $83.3_{\pm 1.5}$ & $87.1_{\pm 1.4}$ & $4.8_{\pm 0.7}$ & $3.9_{\pm 0.5}$ & $80.6_{\pm 1.6}$ & $84.7_{\pm 1.4}$ & $3.1_{\pm 0.5}$ & $6.2_{\pm 0.7}$ \\

Full w/o DR & $77.6_{\pm 1.0}$ & $81.7_{\pm 1.0}$ & $6.4_{\pm 0.7}$ & $5.1_{\pm 0.4}$ & $84.1_{\pm 1.4}$ & $87.6_{\pm 1.3}$ & $4.3_{\pm 0.6}$ & $3.6_{\pm 0.4}$ & $81.1_{\pm 1.5}$ & $85.8_{\pm 1.3}$ & $2.9_{\pm 0.4}$ & $5.5_{\pm 1.0}$ \\

Full  & \bm{$78.8_{\pm 1.2}$} &  \bm{$82.4_{\pm 1.0}$} & \underline{$6.1_{\pm 0.4}$} & \underline{$3.5_{\pm 0.3}$} & \bm{$84.8_{\pm 1.6}$} & \bm{$88.9_{\pm 1.5}$} & \underline{$1.5_{\pm 0.3}$} & \underline{$3.0_{\pm 0.6}$} & \bm{$81.6_{\pm 1.8}$} & \bm{$86.4_{\pm 1.6}$} & \underline{$2.8_{\pm 0.7}$} & \underline{$5.2_{\pm 0.9}$}   \\
\bottomrule
\end{tabular}}
\end{table*}

\end{document}